\pdfoutput=1
\documentclass[runningheads,a4paper]{llncs}

\PassOptionsToPackage{table}{xcolor}
\usepackage[table]{xcolor}

\usepackage{amsfonts}
\usepackage{graphicx}
\usepackage{caption}
\usepackage{subcaption}
\usepackage{enumerate}

\RequirePackage{graphics}
\usepackage{footnote}
\usepackage{xparse}
\usepackage{cite}
\usepackage{booktabs}

\usepackage{booktabs, multirow, makecell}

\usepackage[hyphens]{url}
\usepackage{amssymb}
\usepackage{amsmath}
\usepackage{keyval}
\usepackage{xspace}
\usepackage{paralist}
\usepackage{listings}
\usepackage{multirow}
\usepackage{stmaryrd}
\usepackage{adjustbox}
\usepackage{makecell}
\usepackage{sidecap}
\usepackage{booktabs}
\usepackage{array}
\RequirePackage{tikz}
\usetikzlibrary{arrows,automata,shapes,calc,through,decorations.pathmorphing,decorations.fractals,chains,shapes.multipart}

\usepackage[pdftex]{hyperref}
\hypersetup{
  colorlinks=true,
  citecolor={blue},
  linkcolor={blue},
  urlcolor={blue}
}
\usepackage{arydshln}
\usepackage[free-standing-units]{siunitx}
\usepackage{tabularx}

\def\authorsam{Samuel Sasaki}
\def\authoranne{Anne M. Tumlin}
\def\authordiego{Diego Manzanas Lopez}
\def\authortaylor{Taylor T. Johnson}
\def\authornavid{Navid Hashemi}
\def\authorben{Ben Wooding}
\def\authorhz{Hongchao Zhang}
\def\authorusama{Muhammad Usama Zubair}
\def\authorwaseem{Waseem Abbas}

\def\authoripek{Ipek Oguz}
\def\authormeiyi{Meiyi Ma}

\usepackage{listings}

\newcommand{\commenttaylor}[1]{}

\newcommand{\nnnum}[1]{\relax\ifmmode
  {\mathbb #1}_{\geq 0} \else ${\mathbb #1}_{\geq 0}$
  \fi}
\newcommand{\npnum}[1]{\relax\ifmmode
  {\mathbb #1}_{\leq 0} \else ${\mathbb #1}_{\leq 0}$
  \fi}
\newcommand{\pnum}[1]{\relax\ifmmode
  {\mathbb #1}_{> 0} \else ${\mathbb #1}_{> 0}$
  \fi}
\newcommand{\nnum}[1]{\relax\ifmmode
  {\mathbb #1}_{< 0} \else ${\mathbb #1}_{< 0}$
  \fi}
\newcommand{\plnum}[1]{\relax\ifmmode
  {\mathbb #1}_{+} \else ${\mathbb #1}_{+}$
  \fi}
\newcommand{\nenum}[1]{\relax\ifmmode
  {\mathbb #1}_{-} \else ${\mathbb #1}_{-}$
  \fi}

\newcommand{\extb}[1]{\relax\ifmmode {\sf ExtBeh}_{#1} \else ${\sf ExtBeh}_{#1}$\fi}
\newcommand{\tdists}[1]{\relax\ifmmode {\sf Tdists}_{#1} \else ${\sf Tdists}_{#1}$\fi}

\newcommand{\exec}[1]{\relax\ifmmode {\sf Execs}_{#1} \else ${\sf Exec}_{#1}$\fi}
\newcommand{\execf}[1]{\relax\ifmmode {\sf Execs}^*_{#1} \else ${\sf Exec}^*_{#1}$\fi}
\newcommand{\execi}[1]{\relax\ifmmode {\sf Execs}^\omega_{#1} \else ${\sf Exec}^\omega_{#1}$\fi}

\newcommand{\ctrace}[1]{\relax\ifmmode {\sf Ctraces}_{#1} \else ${\sf Ctraces}_{#1}$\fi}

\newcommand{\trace}[1]{\relax\ifmmode {\sf Traces}_{#1} \else ${\sf Traces}_{#1}$\fi}
\newcommand{\tracef}[1]{\relax\ifmmode {\sf Traces}^*_{#1} \else ${\sf Traces}^*_{#1}$\fi}
\newcommand{\tracei}[1]{\relax\ifmmode {\sf Traces}^\omega_{#1} \else ${\sf Traces}^\omega_{#1}$\fi}

\newcommand{\frag}[1]{\relax\ifmmode {\sf Frags}_{#1} \else ${\sf Frags}_{#1}$\fi}
\newcommand{\fragf}[1]{\relax\ifmmode {\sf Frags}^*_{#1} \else ${\sf Frags}^*_{#1}$\fi}
\newcommand{\fragi}[1]{\relax\ifmmode {\sf Frags}^\omega_{#1} \else ${\sf Frags}^\omega_{#1}$\fi}

\newcommand{\reach}[1]{\relax\ifmmode {\sf Reach}_{#1} \else ${\sf Reach}_{#1}$\fi}

\def\A{{\cal A}}

\def\E{{\cal E}}

\def\I{{\cal I}}

\def\R{{\cal R}}

\def\T{{\cal T}}

\def\U{{\cal U}}

\newcommand{\col}[1]{\relax\ifmmode \mathscr #1\else $\mathscr #1$\fi}

\definecolor{HIOAcolor}{rgb}{0.776,0.22,0.07}

\newcommand{\SC}[2]{\relax\ifmmode {\tt Scount}(#1,#2) \else ${\tt Scount}(#1,#2)$\fi}
\newcommand{\SCM}[2]{\relax\ifmmode {\tt Smin}(#1,#2) \else ${\tt Smin}(#1,#2)$\fi}
\newcommand{\Aut}[1]{\relax\ifmmode {\tt Aut}(#1) \else ${\tt Aut}(#1)$\fi}

\newcommand{\act}[1]{{\operatorname{\mathsf{#1}}}}

\renewcommand{\eqref}[1]{Equation~\ref{eq:#1}}

\newcommand{\remove}[1]{}
\newcommand{\salg}[1]{\relax\ifmmode {\mathcal F}_{#1}\else ${\mathcal F}_{#1}$\fi}
\newcommand{\msp}[1]{\relax\ifmmode (#1, \salg{#1}) \else $(#1, \salg{#1})$\fi}
\newcommand{\msprod}[2]{\relax\ifmmode ( #1 \times #2, \salg{#1} \otimes \salg{#2}) \else $(#1 \times #2, \salg{#1} \otimes \salg{#2})$\fi}
\newcommand{\dist}[1]{\relax\ifmmode {\mathcal P}\msp{#1}
  \else ${\mathcal P}\msp{#1}$\fi}
\newcommand{\subdist}[1]{\relax\ifmmode {\mathcal S}{\mathcal P}\msp{#1}
  \else ${\mathcal S}{\mathcal P}\msp{#1}$\fi}
\newcommand{\disc}[1]{\relax\ifmmode {\sf Disc}(#1)
  \else ${\sf Disc}(#1)$\fi}

\newcommand{\Trajeq}{\relax\ifmmode {\mathcal R}_\T \else ${\mathcal R}_\T$\fi}
\newcommand{\Acteq}{\relax\ifmmode {\mathcal R}_A \else ${\mathcal R}_A$\fi}
\newcommand{\noop}{\relax\ifmmode \lambda \else $\lambda$\fi}
\newcommand{\close}[1]{\relax\ifmmode \overline{#1} \else $\overline{#1}$\fi}

\newcommand{\tup}[1]
           {
             \relax\ifmmode
             \langle #1 \rangle
             \else $\langle$ #1 $\rangle$ \fi
           }

\newcommand{\lit}[1]{ \relax\ifmmode
                \mathord{\mathcode`\-="702D\sf #1\mathcode`\-="2200}
                \else {\it #1} \fi }

\newcommand{\figuresize}{\scriptsize}

\lstdefinelanguage{ioa}{
  basicstyle=\figuresize,
  keywordstyle=\bf \figuresize,
  identifierstyle=\it \figuresize,
  emphstyle=\tt \figuresize,
  mathescape=true,
  tabsize=20,
  sensitive=false,
  columns=fullflexible,
  keepspaces=false,
  flexiblecolumns=true,
  basewidth=0.05em,
  escapeinside={(*@}{@*)},
  moredelim=[il][\rm]{//},
  moredelim=[is][\sf \figuresize]{!}{!},
  moredelim=[is][\bf \figuresize]{*}{*},
  keywords={automaton,and,
  	 choose,const,continue, components,
  	 discrete, do,
  	 eff, Eff, external,else, elseif, evolve, end,
  	 fi,for, forward, from,
  	 hidden,
  	 in,input,internal,if,invariant, initially, imports,
     let,
     or, output, operators, od, of,
     pre, Pre,
     return,
     such,satisfies, stop, signature, simulation,
     trajectories,trajdef, transitions, that,then, type, types, to, tasks,
     variables, vocabulary,
     when,where, with,while},
  emph={set, seq, tuple, map, array, enumeration},
   literate=
        {(}{{$($}}1
        {)}{{$)$}}1
        {\\in}{{$\in\ $}}1
        {\\preceq}{{$\preceq\ $}}1
        {\\subset}{{$\subset\ $}}1
        {\\subseteq}{{$\subseteq\ $}}1
        {\\supset}{{$\supset\ $}}1
        {\\supseteq}{{$\supseteq\ $}}1
        {\\forall}{{$\forall$}}1
        {\\le}{{$\le\ $}}1
        {\\ge}{{$\ge\ $}}1
        {\\gets}{{$\gets\ $}}1
        {\\cup}{{$\cup\ $}}1
        {\\cap}{{$\cap\ $}}1
        {\\langle}{{$\langle$}}1
        {\\rangle}{{$\rangle$}}1
        {\\exists}{{$\exists\ $}}1
        {\\bot}{{$\bot$}}1
        {\\rip}{{$\rip$}}1
        {\\emptyset}{{$\emptyset$}}1
        {\\notin}{{$\notin\ $}}1
        {\\not\\exists}{{$\not\exists\ $}}1
        {\\ne}{{$\ne\ $}}1
        {\\to}{{$\to\ $}}1
        {\\implies}{{$\implies\ $}}1
        {<}{{$<\ $}}1
        {>}{{$>\ $}}1
        {=}{{$=\ $}}1
        {~}{{$\neg\ $}}1
        {|}{{$\mid$}}1
        {'}{{$^\prime$}}1
        {\\A}{{$\forall\ $}}1
        {\\E}{{$\exists\ $}}1
        {\\nE}{{$\nexists\ $}}1
        {\\/}{{$\vee\,$}}1
        {\\vee}{{$\vee\,$}}1
        {/\\}{{$\wedge\,$}}1
        {\\wedge}{{$\wedge\,$}}1
        {=>}{{$\Rightarrow\ $}}1
        {->}{{$\rightarrow\ $}}1
        {<=}{{$\Leftarrow\ $}}1
        {<-}{{$\leftarrow\ $}}1
        {~=}{{$\neq\ $}}1
        {\\U}{{$\cup\ $}}1
        {\\I}{{$\cap\ $}}1
        {|-}{{$\vdash\ $}}1
        {-|}{{$\dashv\ $}}1
        {<<}{{$\ll\ $}}2
        {>>}{{$\gg\ $}}2
        {||}{{$\|$}}1
        {[}{{$[$}}1
        {]}{{$\,]$}}1
        {[[}{{$\langle$}}1
        {]]]}{{$]\rangle$}}1
        {]]}{{$\rangle$}}1
        {<=>}{{$\Leftrightarrow\ $}}2
        {<->}{{$\leftrightarrow\ $}}2
        {(+)}{{$\oplus\ $}}1
        {(-)}{{$\ominus\ $}}1
        {_i}{{$_{i}$}}1
        {_j}{{$_{j}$}}1
        {_{i,j}}{{$_{i,j}$}}3
        {_{j,i}}{{$_{j,i}$}}3
        {_0}{{$_0$}}1
        {_1}{{$_1$}}1
        {_2}{{$_2$}}1
        {_n}{{$_n$}}1
        {_p}{{$_p$}}1
        {_k}{{$_n$}}1
        {-}{{$\ms{-}$}}1
        {@}{{}}0
        {\\delta}{{$\delta$}}1
        {\\R}{{$\R$}}1
        {\\Rplus}{{$\Rplus$}}1
        {\\N}{{$\N$}}1
        {\\times}{{$\times\ $}}1
        {\\tau}{{$\tau$}}1
        {\\alpha}{{$\alpha$}}1
        {\\beta}{{$\beta$}}1
        {\\gamma}{{$\gamma$}}1
        {\\ell}{{$\ell\ $}}1
        {--}{{$-\ $}}1
        {\\TT}{{\hspace{1.5em}}}3
      }

\lstdefinelanguage{ioaNums}[]{ioa}
{
  numbers=left,
  numberstyle=\tiny,
  stepnumber=2,
  numbersep=4pt
}

\lstdefinelanguage{ioaNumsRight}[]{ioa}
{
  numbers=right,
  numberstyle=\tiny,
  stepnumber=2,
  numbersep=4pt
}

\lstnewenvironment{IOA}%
  {\lstset{language=IOA}}
  {}

\lstnewenvironment{IOANums}%
  {
  \if@firstcolumn
    \lstset{language=IOA, numbers=left, firstnumber=auto}
  \else
    \lstset{language=IOA, numbers=right, firstnumber=auto}
  \fi
  }
  {}

\lstnewenvironment{IOANumsRight}%
  {
    \lstset{language=IOA, numbers=right, firstnumber=auto}
  }
  {}

\newcommand{\linefigioa}[9]{

}

\lstdefinelanguage{ioaLang}{%
  basicstyle=\ttfamily\small,
  keywordstyle=\rmfamily\bfseries\small,
  identifierstyle=\small,
  keywords={assumes,automaton,axioms,backward,bounds,by,case,choose,components,const,d,det,discrete,do,eff,else,elseif,ensuring,enumeration,evolve,fi,fire,follow,for,forward,from,hidden,if,in,%
    input,initially,internal,invariant,let, local,od,of,output,pre,schedule,signature,so,%
    simulation,states,variables, tasks, stop,tasks,that,then,to,trajdef,trajectory,trajectories,transitions,tuple,type,union,urgent,uses,when,where,while,yield},
  literate=
        {\\in}{{$\in$}}1
        {\\preceq}{{$\preceq$}}1
        {\\subset}{{$\subset$}}1
        {\\subseteq}{{$\subseteq$}}1
        {\\supset}{{$\supset$}}1
        {\\supseteq}{{$\supseteq$}}1
        {\\rho}{{$\rho$}}1
        {\\infty}{{$\infty$}}1
        {<}{{$<$}}1
        {>}{{$>$}}1
        {=}{{$=$}}1
        {~}{{$\neg$}}1
        {|}{{$\mid$}}1
        {'}{{$^\prime$}}1
        {\\A}{{$\forall$}}1 {\\E}{{$\exists$}}1
        {\\/}{{$\vee$}}1 {/\\}{{$\wedge$}}1
        {=>}{{$\Rightarrow$}}1
        {->}{{$\rightarrow$}}1
        {<=}{{$\leq$}}1 {>=}{{$\geq$}}1 {~=}{{$\neq$}}1
        {\\U}{{$\cup$}}1 {\\I}{{$\cap$}}1
        {|-}{{$\vdash$}}1 {-|}{{$\dashv$}}1
        {<<}{{$\ll$}}2 {>>}{{$\gg$}}2
        {||}{{$\|$}}1
        {<=>}{{$\Leftrightarrow$}}2
        {<->}{{$\leftrightarrow$}}2
        {(+)}{{$\oplus$}}1
        {(-)}{{$\ominus$}}1
}

\lstdefinelanguage{bigIOALang}{%
  basicstyle=\ttfamily,
  keywordstyle=\rmfamily\bfseries,
  identifierstyle=,
  keywords={assumes,automaton,axioms,backward,by,case,choose,components,const,%
    d,det,discrete,do,eff,else,elseif,ensuring,enumeration,evolve,fi,for,forward,from,hidden,if,in%
    input,initially,internal,invariant,local,od,of,output,pre,schedule,signature,so,%
    tasks, simulation,states,stop,tasks,that,then,to,trajdef,trajectories,transitions,tuple,type,union,urgent,uses,when,where,yield},
  literate=
        {\\in}{{$\in$}}1
        {\\preceq}{{$\preceq$}}1
        {\\subset}{{$\subset$}}1
        {\\subseteq}{{$\subseteq$}}1
        {\\supset}{{$\supset$}}1
        {\\supseteq}{{$\supseteq$}}1
        {<}{{$<$}}1
        {>}{{$>$}}1
        {=}{{$=$}}1
        {~}{{$\neg$}}1
        {|}{{$\mid$}}1
        {'}{{$^\prime$}}1
        {\\A}{{$\forall$}}1 {\\E}{{$\exists$}}1
        {\\/}{{$\vee$}}1 {/\\}{{$\wedge$}}1
        {=>}{{$\Rightarrow$}}1
        {->}{{$\rightarrow$}}1
        {<=}{{$\leq$}}1 {>=}{{$\geq$}}1 {~=}{{$\neq$}}1
        {\\U}{{$\cup$}}1 {\\I}{{$\cap$}}1
        {|-}{{$\vdash$}}1 {-|}{{$\dashv$}}1
        {<<}{{$\ll$}}2 {>>}{{$\gg$}}2
        {||}{{$\|$}}1
        {<=>}{{$\Leftrightarrow$}}2
        {<->}{{$\leftrightarrow$}}2
        {(+)}{{$\oplus$}}1
        {(-)}{{$\ominus$}}1
}

\lstnewenvironment{BigIOA}%
  {\lstset{language=bigIOALang,basicstyle=\ttfamily}
   \csname lst@SetFirstLabel\endcsname}
  {\csname lst@SaveFirstLabel\endcsname\vspace{-4pt}\noindent}

\lstnewenvironment{SmallIOA}%
  {\lstset{language=ioaLang,basicstyle=\ttfamily\scriptsize}
   \csname lst@SetFirstLabel\endcsname}
  {\csname lst@SaveFirstLabel\endcsname\noindent}

\newlength{\bracklen}

\renewcommand{\arraystretch}{\defaultArraystretch}

\newcommand{\tri}[3]{\ensuremath{\mathit{#1}^\mathit{#2}_\mathit{#3}}}

\newcommand{\sugLocalVars}[2]{\ifthenelse{\equal{}{#2}}%
                             {\tri{localVars}{#1}{desug}}%
                             {\tri{localVars}{#1}{#2,desug}}}
\newcommand{\sugVars}[2]{\ifthenelse{\equal{}{#2}}%
                        {\tri{vars}{#1}{desug}}%
                        {\tri{vars}{#1}{#2,desug}}}

\newenvironment{subSyntax}{\begin{array}{l}}{\end{array}}

\newcommand{\ms}[1]{\ifmmode%
\mathord{\mathcode`-="702D\it #1\mathcode`\-="2200}\else%
$\mathord{\mathcode`-="702D\it #1\mathcode`\-="2200}$\fi}

\def\A{{\cal A}}

\def\T{{\cal T}}

\lstdefinelanguage{pvs}{
  basicstyle=\tt \figuresize,
  keywordstyle=\sc \figuresize,
  identifierstyle=\it \figuresize,
  emphstyle=\tt \figuresize,
  mathescape=true,
  tabsize=20,
  sensitive=false,
  columns=fullflexible,
  keepspaces=false,
  flexiblecolumns=true,
  basewidth=0.05em,
  moredelim=[il][\rm]{//},
  moredelim=[is][\sf \figuresize]{!}{!},
  moredelim=[is][\bf \figuresize]{*}{*},
  keywords={and,
  	 begin,
  	 cases, const,
  	 do,
  	 external, else, exists, end, endcases, endif,
  	 fi,for, forall, from,
  	 hidden,
  	 in, if, importing,
     let, lambda, lemma,
     measure,
     not,
     or, of,
     return, recursive,
     stop,
     theory, that,then, type, types, type+, to, theorem,
     var,
     with,while},
  emph={nat, setof, sequence, eq, tuple, map, array, enumeration, bool, real, exp, nnreal, posreal},
   literate=
        {(}{{$($}}1
        {)}{{$)$}}1
        {\\in}{{$\in\ $}}1
        {\\mapsto}{{$\rightarrow\ $}}1
        {\\preceq}{{$\preceq\ $}}1
        {\\subset}{{$\subset\ $}}1
        {\\subseteq}{{$\subseteq\ $}}1
        {\\supset}{{$\supset\ $}}1
        {\\supseteq}{{$\supseteq\ $}}1
        {\\forall}{{$\forall$}}1
        {\\le}{{$\le\ $}}1
        {\\ge}{{$\ge\ $}}1
        {\\gets}{{$\gets\ $}}1
        {\\cup}{{$\cup\ $}}1
        {\\cap}{{$\cap\ $}}1
        {\\langle}{{$\langle$}}1
        {\\rangle}{{$\rangle$}}1
        {\\exists}{{$\exists\ $}}1
        {\\bot}{{$\bot$}}1
        {\\rip}{{$\rip$}}1
        {\\emptyset}{{$\emptyset$}}1
        {\\notin}{{$\notin\ $}}1
        {\\not\\exists}{{$\not\exists\ $}}1
        {\\ne}{{$\ne\ $}}1
        {\\to}{{$\to\ $}}1
        {\\implies}{{$\implies\ $}}1
        {<}{{$<\ $}}1
        {>}{{$>\ $}}1
        {=}{{$=\ $}}1
        {~}{{$\neg\ $}}1
        {|}{{$\mid$}}1
        {'}{{$^\prime$}}1
        {\\A}{{$\forall\ $}}1
        {\\E}{{$\exists\ $}}1
        {\\/}{{$\vee\,$}}1
        {\\vee}{{$\vee\,$}}1
        {/\\}{{$\wedge\,$}}1
        {\\wedge}{{$\wedge\,$}}1
        {->}{{$\rightarrow\ $}}1
        {=>}{{$\Rightarrow\ $}}1
        {->}{{$\rightarrow\ $}}1
        {<=}{{$\Leftarrow\ $}}1
        {<-}{{$\leftarrow\ $}}1
        {~=}{{$\neq\ $}}1
        {\\U}{{$\cup\ $}}1
        {\\I}{{$\cap\ $}}1
        {|-}{{$\vdash\ $}}1
        {-|}{{$\dashv\ $}}1
        {<<}{{$\ll\ $}}2
        {>>}{{$\gg\ $}}2
        {||}{{$\|$}}1
        {[}{{$[$}}1
        {]}{{$\,]$}}1
        {[[}{{$\langle$}}1
        {]]]}{{$]\rangle$}}1
        {]]}{{$\rangle$}}1
        {<=>}{{$\Leftrightarrow\ $}}2
        {<->}{{$\leftrightarrow\ $}}2
        {(+)}{{$\oplus\ $}}1
        {(-)}{{$\ominus\ $}}1
        {_i}{{$_{i}$}}1
        {_j}{{$_{j}$}}1
        {_{i,j}}{{$_{i,j}$}}3
        {_{j,i}}{{$_{j,i}$}}3
        {_0}{{$_0$}}1
        {_1}{{$_1$}}1
        {_2}{{$_2$}}1
        {_n}{{$_n$}}1
        {_p}{{$_p$}}1
        {_k}{{$_n$}}1
        {-}{{$\ms{-}$}}1
        {@}{{}}0
        {\\delta}{{$\delta$}}1
        {\\R}{{$\R$}}1
        {\\Rplus}{{$\Rplus$}}1
        {\\N}{{$\N$}}1
        {\\times}{{$\times\ $}}1
        {\\tau}{{$\tau$}}1
        {\\alpha}{{$\alpha$}}1
        {\\beta}{{$\beta$}}1
        {\\gamma}{{$\gamma$}}1
        {\\ell}{{$\ell\ $}}1
        {--}{{$-\ $}}1
        {\\TT}{{\hspace{1.5em}}}3
      }

\lstdefinelanguage{BigPVS}{
  basicstyle=\tt,
  keywordstyle=\sc,
  identifierstyle=\it,
  emphstyle=\tt ,
  mathescape=true,
  tabsize=20,
  sensitive=false,
  columns=fullflexible,
  keepspaces=false,
  flexiblecolumns=true,
  basewidth=0.05em,
  moredelim=[il][\rm]{//},
  moredelim=[is][\sf \figuresize]{!}{!},
  moredelim=[is][\bf \figuresize]{*}{*},
  keywords={and,
  	 begin,
  	 cases, const,
  	 do, datatype,
  	 external, else, exists, end, endif, endcases,
  	 fi,for, forall, from,
  	 hidden,
  	 in, if, importing,
     let, lambda, lemma,
     measure,
     not,
     or, of,
     return, recursive,
     stop,
     theory, that,then, type, types, type+, to, theorem,
     var,
     with,while},
  emph={nat, setof, sequence, eq, tuple, map, array, first, rest, add, enumeration, bool, real, posreal, nnreal},
   literate=
        {(}{{$($}}1
        {)}{{$)$}}1
        {\\in}{{$\in\ $}}1
        {\\mapsto}{{$\rightarrow\ $}}1
        {\\preceq}{{$\preceq\ $}}1
        {\\subset}{{$\subset\ $}}1
        {\\subseteq}{{$\subseteq\ $}}1
        {\\supset}{{$\supset\ $}}1
        {\\supseteq}{{$\supseteq\ $}}1
        {\\forall}{{$\forall$}}1
        {\\le}{{$\le\ $}}1
        {\\ge}{{$\ge\ $}}1
        {\\gets}{{$\gets\ $}}1
        {\\cup}{{$\cup\ $}}1
        {\\cap}{{$\cap\ $}}1
        {\\langle}{{$\langle$}}1
        {\\rangle}{{$\rangle$}}1
        {\\exists}{{$\exists\ $}}1
        {\\bot}{{$\bot$}}1
        {\\rip}{{$\rip$}}1
        {\\emptyset}{{$\emptyset$}}1
        {\\notin}{{$\notin\ $}}1
        {\\not\\exists}{{$\not\exists\ $}}1
        {\\ne}{{$\ne\ $}}1
        {\\to}{{$\to\ $}}1
        {\\implies}{{$\implies\ $}}1
        {<}{{$<\ $}}1
        {>}{{$>\ $}}1
        {=}{{$=\ $}}1
        {~}{{$\neg\ $}}1
        {|}{{$\mid$}}1
        {'}{{$^\prime$}}1
        {\\A}{{$\forall\ $}}1
        {\\E}{{$\exists\ $}}1
        {\\/}{{$\vee\,$}}1
        {\\vee}{{$\vee\,$}}1
        {/\\}{{$\wedge\,$}}1
        {\\wedge}{{$\wedge\,$}}1
        {->}{{$\rightarrow\ $}}1
        {=>}{{$\Rightarrow\ $}}1
        {->}{{$\rightarrow\ $}}1
        {<=}{{$\Leftarrow\ $}}1
        {<-}{{$\leftarrow\ $}}1
        {~=}{{$\neq\ $}}1
        {\\U}{{$\cup\ $}}1
        {\\I}{{$\cap\ $}}1
        {|-}{{$\vdash\ $}}1
        {-|}{{$\dashv\ $}}1
        {<<}{{$\ll\ $}}2
        {>>}{{$\gg\ $}}2
        {||}{{$\|$}}1
        {[}{{$[$}}1
        {]}{{$\,]$}}1
        {[[}{{$\langle$}}1
        {]]]}{{$]\rangle$}}1
        {]]}{{$\rangle$}}1
        {<=>}{{$\Leftrightarrow\ $}}2
        {<->}{{$\leftrightarrow\ $}}2
        {(+)}{{$\oplus\ $}}1
        {(-)}{{$\ominus\ $}}1
        {_i}{{$_{i}$}}1
        {_j}{{$_{j}$}}1
        {_{i,j}}{{$_{i,j}$}}3
        {_{j,i}}{{$_{j,i}$}}3
        {_0}{{$_0$}}1
        {_1}{{$_1$}}1
        {_2}{{$_2$}}1
        {_n}{{$_n$}}1
        {_p}{{$_p$}}1
        {_k}{{$_n$}}1
        {-}{{$\ms{-}$}}1
        {@}{{}}0
        {\\delta}{{$\delta$}}1
        {\\R}{{$\R$}}1
        {\\Rplus}{{$\Rplus$}}1
        {\\N}{{$\N$}}1
        {\\times}{{$\times\ $}}1
        {\\tau}{{$\tau$}}1
        {\\alpha}{{$\alpha$}}1
        {\\beta}{{$\beta$}}1
        {\\gamma}{{$\gamma$}}1
        {\\ell}{{$\ell\ $}}1
        {--}{{$-\ $}}1
        {\\TT}{{\hspace{1.5em}}}3
      }

\lstdefinelanguage{pvsNums}[]{pvs}
{
  numbers=left,
  numberstyle=\tiny,
  stepnumber=2,
  numbersep=4pt
}

\lstdefinelanguage{pvsNumsRight}[]{pvs}
{
  numbers=right,
  numberstyle=\tiny,
  stepnumber=2,
  numbersep=4pt
}

\lstnewenvironment{BigPVS}%
  {\lstset{language=BigPVS}}
  {}

\lstnewenvironment{PVSNums}%
  {
  \if@firstcolumn
    \lstset{language=pvs, numbers=left, firstnumber=auto}
  \else
    \lstset{language=pvs, numbers=right, firstnumber=auto}
  \fi
  }
  {}

\lstnewenvironment{PVSNumsRight}%
  {
    \lstset{language=pvs, numbers=right, firstnumber=auto}
  }
  {}

\newcommand{\linefigpvs}[9]{

}

\lstdefinelanguage{pvsproof}{
  basicstyle=\tt \figuresize,
  mathescape=true,
  tabsize=4,
  sensitive=false,
  columns=fullflexible,
  keepspaces=false,
  flexiblecolumns=true,
  basewidth=0.05em,
}

\def\N{\act{N}}

\newcommand{\localvar}[2]{{{#1_{#2}}}}

\def\xi{\localvar{x}{i}}

\def\reach{{\sf Reach}}

\def\Xi{\mathit{X_i}}

\begin{document}

\title{\textsf{NNV3}: Expanding Neural Network Verification to New Architectures and Domains\thanks{A.Tumlin and S.Sasaki are co-first authors.}}

\titlerunning{\textsf{NNV3}: The Neural Network Verification Tool}

\author{\hspace{0.1cm} \authoranne \inst{1} \and \authorsam \inst{1} \and \authorben \inst{1} \and \authordiego \inst{1} \and  \authorusama \inst{2} \and \authornavid \inst{1} \and \authorhz \inst{1}  \and \authorwaseem \inst{2} \and \authoripek \inst{1} \and \authormeiyi \inst{1} \and \authortaylor \inst{1}}

\authorrunning{Tumlin, Sasaki, et al.}

\institute{Vanderbilt University, USA \and The University of Texas at Dallas, USA}

\maketitle

\setcounter{footnote}{0}

\begin{abstract}
We present \textsf{NNV3}, the latest version of the Neural Network Verification (\textsf{NNV}) tool, a \textsf{MATLAB} framework for formal verification of deep learning models and learning-enabled cyber-physical systems. Building on the set-based reachability foundation of \textsf{NNV} 1.0 (FFNNs, CNNs, NNCS) and \textsf{NNV} 2.0 (RNNs, SSNNs, neural ODEs), \textsf{NNV3} introduces new members of the Star-set family: ModelStar for verifying networks under weight perturbation, VolumeStar for video and 3D volumetric inputs, and GraphStar for graph neural networks. A conformal-inference-based probabilistic reachability mode complements sound analysis for problems where deterministic verification is intractable, while FairNNV certifies counterfactual and individual fairness properties over continuous input regions. \textsf{NNV3} introduces new benchmarks for malware detection, graph-based power-system models, medical imaging, variable-length time series data, and action recognition. \textsf{NNV3} also incorporates tutorials and developer guides through a unified documentation site. This paper details these major updates, demonstrating \textsf{NNV}'s maturation into a comprehensive, robust, and accessible verification tool for a diverse range of AI systems.

\end{abstract}

\section{Introduction}
Deep neural networks (DNNs) have become integral to solving complex problems across various domains, from image classification to autonomous control. However, their deployment in safety-critical applications is hindered by their opaque nature and susceptibility to adversarial perturbations. Formal verification provides a means to analyze and rigorously guarantee the behavior of these models, which is essential for establishing trust in AI-powered systems.

The Neural Network Verification (\textsf{NNV})\footnote{\url{https://github.com/verivital/nnv/}} tool~\cite{tran2020cavtool} was introduced as a comprehensive, open-source \textsf{MATLAB} toolbox to tackle this challenge. \textsf{NNV} is built around a powerful computation engine that performs set-based reachability analysis, computing the set of all possible outputs for a given set of inputs. The initial release of \textsf{NNV} focused on providing exact and over-approximate reachability for feed-forward neural networks (FFNNs), convolutional neural networks (CNNs), and neural network control systems (NNCS) using a variety of set representations like polyhedra, zonotopes, and the novel star set.

Building on this foundation, \textsf{NNV} 2.0~\cite{lopez2023cavtool} expanded the tool's scope to handle a wider array of complex and dynamic network architectures. It introduced verification support for neural ordinary differential equations (neural ODEs), recurrent neural networks (RNNs), and semantic segmentation neural networks (SSNNs). This version also improved scalability with new relaxed reachability methods~\cite{tran2021cav} and enhanced usability by supporting standard community formats like \textsf{ONNX}~\cite{onnx} and \textsf{VNNLIB}~\cite{vnnlib}.

This paper presents \textsf{NNV3}, an evolution of the \textsf{NNV} tool that addresses emerging challenges in AI verification and broadens the tool's applicability to new domains and data types. While previous versions focused on expanding architectural support, \textsf{NNV3} introduces verification techniques for new classes of properties and extends reachability analysis to previously intractable data modalities. The contributions span three categories. \textbf{Algorithmic and model improvements} introduce reachability-based verification for parameter perturbations (ModelStar~\cite{zubair2025verification,zubair2026modelstar}), spatio-temporal data (VolumeStar~\cite{sasaki2025videoclass}), graph-structured models (GNNV with GraphStar~\cite{tumlin2026reachability}), probabilistic guarantees via conformal inference~\cite{hashemi2025scaling}, fairness properties (FairNNV~\cite{tumlin2024fairnnv}), and variable-length time-dependent networks~\cite{pal2023fmics}. \textbf{New application domains} unlocked by these algorithms include video classification~\cite{sasaki2025videoclass}, medical-image segmentation~\cite{hashemi2025probabilistic}, power-system analysis~\cite{tumlin2026reachability}, malware detection~\cite{preston2024malware}, ethical decision-making~\cite{tumlin2024fairnnv}, and deployment-uncertainty robustness~\cite{zubair2025verification,zubair2026modelstar}. \textbf{System upgrades} include a unified documentation site\footnote{\url{https://verivital.github.io/nnv/}} that consolidates the user guide, developer guide, API reference, \emph{etc.} These advancements solidify \textsf{NNV}'s position as one of the most comprehensive and versatile verification frameworks available to the research community.

\vspace{-0.3cm}
\section{\textsf{NNV3} vs. \textsf{NNV} 2.0}
\vspace{-0.1cm}

\begin{figure}[t]
    \centering
    \includegraphics[width=\linewidth]{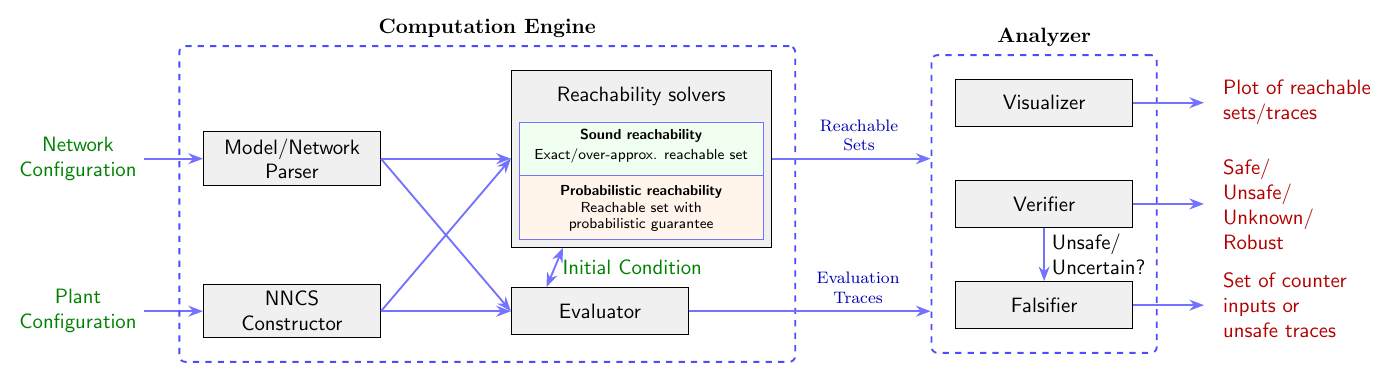}
    \caption{Updated \textsf{NNV} verification pipeline, composed of a computation engine and analyzer, and extended to support both sound and probabilistic reachability analysis.}
    \label{fig:nnv3_engine}
\end{figure}

The core architecture of \textsf{NNV}, illustrated in Fig.~\ref{fig:nnv3_engine}, is composed of two primary modules: the \textbf{Computation Engine} and the \textbf{Analyzer}. The \textbf{Computation Engine} parses neural network and system models (\textsf{ONNX} and \textsf{MATLAB} formats) and performs layer-by-layer reachability analysis using set-based abstractions, including Star sets~\cite{tran2019fm}, ImageStar~\cite{tran2020cav}, and newly introduced representations such as VolumeStar (\emph{a.k.a} VideoStar)~\cite{sasaki2025videoclass}, ModelStar~\cite{zubair2025verification, zubair2026modelstar}, and GraphStar~\cite{tumlin2026reachability}. In \textsf{NNV3}, the computation engine supports both sound reachability and a probabilistic reachability mode~\cite{hashemi2025scaling}, the latter enabling scalable verification via sampling-based uncertainty quantification. The \textbf{Analyzer} consumes the resulting reachable sets or evaluation traces to verify system-level properties --- including robustness, safety specifications expressed in \textsf{VNNLIB}, and fairness constraints --- and supports visualization, verification, and counterexample generation. Together, these extensions broaden the scope of verifiable models while preserving the modular structure of the \textsf{NNV} framework.

Practitioners working with heterogeneous, real-world AI systems require a unified and extensible verification framework. This drives the continued evolution of \textsf{NNV}. Table~\ref{tab:nnv-features} illustrates that \textsf{NNV} and its prior iterations support a broad spectrum of architectures and applications, including feedforward and convolutional networks (FFNNs and CNNs), recurrent models (RNNs), semantic segmentation (SSNNs), neural ODEs, and neural network control systems (NNCSs).  With the release of \textsf{NNV3}, we expand the tool to emerging and increasingly important model classes, including graph neural networks (GNNs) and video classification architectures (VolumeStar). \textsf{NNV3} addresses a broader class of deployment-relevant properties, including algorithmic fairness, parameter and weight perturbations, and probabilistic guarantees. These capabilities enable verification beyond conventional robustness analysis and reflect practical concerns encountered in real-world deployment.

\textsf{NNV} has been evaluated consistently in community benchmarks such as VNN-COMP~\cite{vnncomp2025} and ARCH-COMP~\cite{ARCH_COMP25_Category_Report,sasaki2026arch}, and has served as the foundation for tutorials at DESTION~\cite{tran2020neural}, EMSOFT~\cite{tran2023tutorial}, DSN~\cite{johnson2024tutorial}, SPIE, IAVVC, AAAI, \emph{etc.}, underscoring its maturity and continued relevance to the verification community (see the user guide\footnote{\url{https://verivital.github.io/nnv/user-guide/index.html}} and the tutorials index\footnote{\url{https://verivital.github.io/nnv/examples/index.html}} for further details).

\textbf{Relation to prior publications.} The set representations and analysis modes incorporated into \textsf{NNV3} were introduced individually in prior work~\cite{zubair2026modelstar,sasaki2025videoclass,tumlin2026reachability,hashemi2025scaling,tumlin2024fairnnv}. This work contributes their systematic integration into a unified verification framework. Previously, these capabilities were implemented as independent prototypes with distinct interfaces, dispatch mechanisms, and evaluation pipelines. \textsf{NNV3} consolidates these developments through a common Star-set abstraction, a shared specification interface, and a documented developer API, enabling verification specifications and reachability procedures to be used consistently across supported representations. The resulting framework is further supported by continuous integration testing, unified documentation for the full representation family, and reproducibility packages for each experiment reported in Section~\ref{sec:evaluation}.

The remainder of the paper is organized as follows. Section \ref{sec:overview} summarizes major new features and capabilities. Section \ref{sec:new_domains} introduces the extended domain applications. We then validate new features in Section \ref{sec:evaluation} and refresh the comparison with the MathWorks AI Verification Library. Finally, we discuss related works in Section \ref{sec:related_work}, and conclude the paper in Section \ref{sec:conclusion}.

\rowcolors{2}{gray!15}{white}
\begin{table}[t]
  \centering
  \caption{Overview of major features available in NNV.
  Items in regular weight are NNV~1.0 baseline;
  \textcolor{blue}{\textit{blue italics}} denote additions introduced in NNV~2.0;
  \textcolor{purple}{\textbf{purple bold}} denote new capabilities introduced in NNV3.}
  \label{tab:nnv-features}
  \resizebox{1.0\linewidth}{!}{
  \setlength{\arrayrulewidth}{.1em}
  \renewcommand{\arraystretch}{1.3}
  \begin{tabular}{l|l}
  \hline
  \textbf{Feature} &
  \textbf{Supported}
  (NNV 1.0, \textcolor{blue}{\textit{NNV 2.0}}, \textcolor{purple}{\textbf{NNV3}}) \\ \hline

  Neural Network Type &
  FFNN, CNN,
  \textcolor{blue}{\textit{NeuralODE}},
  \textcolor{blue}{\textit{SSNN}},
  \textcolor{blue}{\textit{RNN}},
  \textcolor{purple}{\textbf{GNN}},
  \textcolor{purple}{\textbf{TDNN}}, \textcolor{purple}{\textbf{3D CNN}} \\

  Layers &
  MaxPool, Conv, BN, AvgPool, FC,
  \textcolor{blue}{\textit{MaxUnpool}},
  \textcolor{blue}{\textit{TC}},
  \textcolor{blue}{\textit{DC}},
  \textcolor{blue}{\textit{NODE}},
  \textcolor{purple}{\textbf{GCN}},
  \textcolor{purple}{\textbf{GINE}}, \textcolor{purple}{\textbf{Conv3D}}\\

  Activation functions &
  ReLU, Satlin, Sigmoid, Tanh,
  \textcolor{blue}{\textit{Leaky ReLU}},
  \textcolor{blue}{\textit{Satlins}} \\

  Plant dynamics (NNCS) &
  Linear ODE, Nonlinear ODE,
  Continuous \& Discrete Time, \textcolor{blue}{\textit{HA}} \\

  Set Representation &
  Polyhedron, Zonotope, Star, ImageStar,
  \textcolor{purple}{\textbf{VolumeStar}},
  \textcolor{purple}{\textbf{ModelStar}},
  \textcolor{purple}{\textbf{GraphStar}} \\

  Star Reach methods &
  exact, approx, abs-dom,
  \textcolor{blue}{\textit{relax-*}} \\

  Reachable set visualization &
  exact and over-approximation \\

  Verification &
  Safety, Robustness,
  \textcolor{blue}{\textit{VNNLIB}},
  \textcolor{purple}{\textbf{Fairness}}, \textcolor{purple}{\textbf{Weight Perturbation}}, \textcolor{purple}{\textbf{Probabilistic}} \\

  Miscellaneous &
  Parallel computing, counterexample generation,
  \textcolor{blue}{\textit{ONNX}}, \textcolor{purple}{\textbf{CI/CD}} \\

  \hline
  \end{tabular}}
  \vspace{-0.5cm}
\end{table}
\rowcolors{1}{white}{white}

\vspace{-0.3cm}
\section{Overview and Features}
\label{sec:overview}
\vspace{-0.1cm}
This section summarizes the major features introduced in \textsf{NNV3} and in Table~\ref{tab:nnv-features}. These additions extend the reachability-based verification capabilities of prior versions of \textsf{NNV} to new model classes, data modalities, and verification objectives, while preserving soundness guarantees and compatibility with existing analysis pipelines. Section~\ref{subsec:feature_highlights} presents the per-feature highlights; Section~\ref{subsec:core_upgrades} describes the underlying engine and extensibility upgrades.

\subsection{Feature Highlights}
\label{subsec:feature_highlights}

\textbf{ModelStar~\cite{zubair2025verification, zubair2026modelstar}:} Neural networks deployed in practice are subject to parameter uncertainty arising from quantization, numerical imprecision, and hardware faults~\cite{sun2021exploring,yan2022computing,guo2018survey}. To address this, \textsf{NNV3} introduces ModelStar, a star-set–based representation for interval-bounded weight perturbations. ModelStar enables reachability analysis under simultaneous input and parameter uncertainty. For networks with a single perturbed layer and singleton inputs, the reachable set of the perturbed layer is computed exactly; for multi-layer perturbations, sound over-approximations are constructed using Star sets.
ModelStar is currently implemented for fully-connected and 2D convolutional layers and can be readily implemented for verification against weight perturbations in any linear layer.
Rounding errors introduced by quantized compression can be modeled as interval-bounded specifications for ModelStar --- as demonstrated in the experiments in Section~\ref{sec:evaluation} --- but fixed-point inference arithmetic (discrete operations) is not yet modeled.\footnote{\url{https://verivital.github.io/nnv/theory/weight-perturbation.html}}

\textbf{VolumeStar (VideoStar)~\cite{sasaki2025videoclass}:}
\textsf{NNV} can formally verify the robustness of video classifiers. Verification of neural networks operating on spatio-temporal data, such as videos and volumetric medical images, presents significant scalability challenges due to input dimensionality. \textsf{NNV3} extends the Star and ImageStar representations~\cite{tran2019fm,tran2020cav} to VolumeStar, a set abstraction for spatio-temporal inputs, \emph{e.g.}, an image at each time step. VolumeStar supports reachability analysis of architectures with 3D convolutional and pooling layers, including video classification networks (such as C3D~\cite{tran2015iccv} and I3D~\cite{carreira2017i3d}), as well as 3D medical imaging models~\cite{yousef2022holistic}. Using VolumeStar reachability, \textsf{NNV} can formally certify classification robustness for all admissible spatio-temporal perturbations within a specified input set. In addition, \textsf{NNV3} extends star-based reachability analysis to time-dependent neural networks (TDNNs) by allowing the analysis horizon to vary over a bounded temporal range, generalizing prior support for fixed-length recurrent models.\footnote{\url{https://verivital.github.io/nnv/theory/imagestar-volumestar.html}}

\textbf{GraphStar~\cite{tumlin2026reachability}:}
Graph neural networks (GNNs) are increasingly used as surrogates for numerical solvers in domains such as molecular modeling, traffic forecasting, and power-system analysis~\cite{wu2021gnnsurvey}. \textsf{NNV3} introduces GNNV, extending reachability-based verification to graph-structured learning models. GNNV implements GraphStar, a generalization of Star sets that captures both graph connectivity and uncertainty in node and edge features. This abstraction enables exact propagation of affine message-passing operations and sound over-approximation of ReLU nonlinearities in common GNN architectures, including graph convolutional networks (GCNs)~\cite{kipf2017semisupervised} and graph isomorphism networks with edge features (GINE)~\cite{hu2020strategies}. GraphStar reachability is demonstrated on power-system case studies, including power flow, optimal power flow, and cascading failure analysis~\cite{varbella2024powergraph}.\footnote{\url{https://verivital.github.io/nnv/theory/gnn-reachability.html}}

\textbf{Probabilistic Verification~\cite{hashemi2025scaling}:}
Exact reachability analysis can become intractable for large networks due to exponential complexity in the number of unstable nonlinear activations. To address this limitation, \textsf{NNV3} integrates a probabilistic, model-agnostic verification approach based on conformal inference. Given an input set, a neural network, and a specification, verification is performed via sampling rather than exhaustive propagation. This approach scales with inference cost and is independent of network architecture, providing probabilistic coverage guarantees for models that are beyond the practical reach of exact methods.\footnote{\url{https://verivital.github.io/nnv/theory/probabilistic.html}}

\textbf{Fairness Verification~\cite{tumlin2024fairnnv}:}
As machine learning systems are deployed in high-stakes decision-making settings, verification of fairness properties has become increasingly important~\cite{mehrabi2021survery}. \textsf{NNV3} integrates FairNNV, a reachability-based framework for formally validating fairness specifications over continuous input regions. FairNNV verifies specifications corresponding to counterfactual fairness~\cite{kusner2017counterfactual}, which requires predictions to remain invariant under changes to sensitive attributes, and individual fairness, which enforces similar outcomes for inputs within a bounded neighborhood. Fairness is quantified using the Verified Fairness (VF) score, defined as the proportion of inputs for which fairness properties are formally certified.\footnote{\url{https://verivital.github.io/nnv/theory/fairness.html}}

\subsection{Core Upgrades \& Extensibility}
\label{subsec:core_upgrades}

\textbf{Star-set family unification.} VolumeStar, GraphStar, and ModelStar are not disjoint abstractions but instances of a common Star-set template, parameterized by an anchor, a generator basis, and a polyhedral constraint on the generator coefficients.

\begin{definition}[Star-set template]
\label{def:star-template}
A \emph{star set} over a tensor space $\mathcal{T}$ is a tuple $\langle c, V, P, q \rangle$ consisting of an \emph{anchor} $c \in \mathcal{T}$, a \emph{generator basis} $V = \{v_1,\dots,v_m\} \subset \mathcal{T}$, and a polyhedral constraint $(P,q) \in \mathbb{R}^{p \times m} \times \mathbb{R}^{p}$ on the predicate variables $\alpha = [\alpha_1,\dots,\alpha_m]^\top$. It denotes the set
\[
\Theta = \Big\{\, x \in \mathcal{T} \;\Big|\; x = c + \textstyle\sum_{i=1}^{m} \alpha_i v_i, \; P\alpha \le q \,\Big\}.
\]
\end{definition}

Each member of the family instantiates Definition~\ref{def:star-template} by fixing $\mathcal{T}$ and the semantics of the generators, and inherits the shared propagation kernel unchanged. VolumeStar, for example, takes $\mathcal{T} = \mathbb{R}^{H \times W \times C \times F}$ for a volume of height $H$, width $W$, $C$ channels, and $F$ frames: the anchor is the nominal video, each generator is a volume encoding one direction of admissible spatio-temporal variation, and $P\alpha \le q$ bounds the perturbation, \emph{e.g.}, $|\alpha_i| \le \epsilon$ for an $L_\infty$ bound applied across all frames. Star and ImageStar differ from it only in the rank of $\mathcal{T}$ ($\mathbb{R}^{n}$ and $\mathbb{R}^{H \times W \times C}$, respectively), whereas GraphStar carries the adjacency structure alongside the node- and edge-feature tensors, and ModelStar draws its generators from perturbed weight entries rather than from input dimensions. Extending \textsf{NNV} to a new modality therefore reduces to instantiating Definition~\ref{def:star-template} and implementing layer-specific dispatch over the shared propagation kernel.

\textbf{Engine-level changes.} Several core-engine upgrades support the new abstractions: a GNN wrapper that maps message-passing layers onto Star-set affine operations and routes adjacency through GraphStar; a 4D-tensor extension of the ImageStar reach kernels that lets VolumeStar inherit existing 3D-conv and pooling implementations; perturbation tracking in ModelStar that maintains symbolic dependencies between input generators and weight-perturbation generators across linear layers; and a probabilistic reachability mode that composes with sound reachability so that any feature implemented for the sound mode is automatically usable in the probabilistic mode.

\textbf{Continuous Integration and Deployment (CI/CD).}
\label{subsec:cicd}
To improve reliability and maintainability, \textsf{NNV3} adopts a continuous integration and continuous deployment pipeline based on GitHub Actions. The pipeline automatically builds and tests the codebase for each commit and pull request, including unit tests for core reachability methods and regression tests on established benchmarks. This ensures backward compatibility and supports sustained development of \textsf{NNV} as an open-source verification tool.

\textbf{Extensibility and developer API.} \textsf{NNV3}'s engine and analyzer modules are documented as a stable developer-facing API\footnote{\url{https://verivital.github.io/nnv/api/index.html}} with guidance for adding new layer types, set representations, and verification properties.\footnote{\url{https://verivital.github.io/nnv/developer/index.html}} The unified documentation site now provides the user guide, developer guide, API reference, theory background, and the tutorials index as a single entry point. Together, these resources lower the barrier for contributors extending \textsf{NNV} to architectures and properties beyond those listed in Table~\ref{tab:nnv-features}.

\vspace{-0.3cm}
\section{New Domains}
\label{sec:new_domains}
\vspace{-0.1cm}
\textsf{NNV3}'s new abstractions and probabilistic mode unlock application domains previously out of reach for reachability-based methods, characterized by high-dimensional inputs, structured representations, or deployment-driven threat models that exceed traditional $L_p$-bounded robustness analysis. Below we highlight four such domains.

\textbf{Malware Detection~\cite{preston2024malware}:}
DNN malware classifiers are vulnerable to adversarial evasion, where attackers apply small functionality-preserving modifications to bypass detection. \textsf{NNV3} introduces a benchmark for verifying robustness of static-feature classifiers under realistic feature-space perturbations (e.g., modifying non-executable sections, appending benign bytes), enabling formal assessment of classifier resilience under a well-defined threat model.

\textbf{Power-System Analysis~\cite{tumlin2026reachability}:}
Power flow, optimal power flow, and cascading failure analysis are naturally graph-structured and increasingly use GNN surrogates~\cite{varbella2024powergraph}, yet prediction errors carry severe operational consequences. Through GNN verification, \textsf{NNV3} provides the first general-purpose framework supporting reachability analysis of such topology-aware models with both node- and edge-feature uncertainty.

\textbf{Medical Imaging Classification~\cite{hashemi2025scaling,hashemi2025probabilistic}:}
High-dimensional semantic segmentation defeats exact verification. \textsf{NNV3}'s probabilistic pipeline analyzes large segmentation networks on lung X-ray datasets~\cite{jaeger2014lung,candemir2014lung} with dense pixel-level outputs, providing coverage guarantees while substantially reducing conservatism relative to exact methods~\cite{hashemi2025scaling}.

\textbf{Financial Predictions~\cite{tumlin2024fairnnv}:}
FairNNV extends \textsf{NNV3} to consequential decision making systems such as credit approval and loan risk assessment~\cite{adult_census,german_credit,bank_marketing}. Unlike statistical auditing over finite datasets, reachability-based fairness analysis certifies properties over continuous input regions, formally reasoning about bias under input perturbations and counterfactual scenarios.

\vspace{-0.3cm}
\section{Evaluation}
\label{sec:evaluation}
\vspace{-0.1cm}
\textsf{NNV3}'s primary contribution is its breadth of supported architectures, specifications, and threat models, many of which lack a direct comparison, as summarized in Table~\ref{tab:tool_comparison}. Accordingly, the evaluations presented here are intended as feasibility demonstrations of the integrated framework across domains rather than exhaustive scalability studies of individual features. The per-feature suites are intentionally compact, allowing the full evaluation to be reproduced within several hours while exercising each supported analysis pipeline. More extensive feature-specific comparisons are reported in the corresponding prior works, including GraphStar against CORA~\cite{tumlin2026reachability} and the conformal verification pipeline~\cite{hashemi2025scaling}. We summarize the relevant results below and additionally refresh the NNV~2.0 head-to-head evaluation~\cite{lopez2023cavtool} against the MathWorks AI Verification Library (AIVL)~\cite{matlabDNNverification} for FFNN and CNN models supported by both tools. All experiments were conducted using the MATLAB 2025b Docker container on a machine equipped with an \texttt{Intel 24 Core i9-285K} CPU, 64~GB RAM, and an \texttt{NVIDIA RTX 5090} GPU with 32~GB VRAM.\footnote{The evaluation artifact (v3.0-atva26) is archived at \url{https://doi.org/10.5281/zenodo.20433720}; the live repository is available at \url{https://github.com/verivital/nnv}.}

\begin{figure}[t]
\centering
\includegraphics[width=0.8\linewidth]{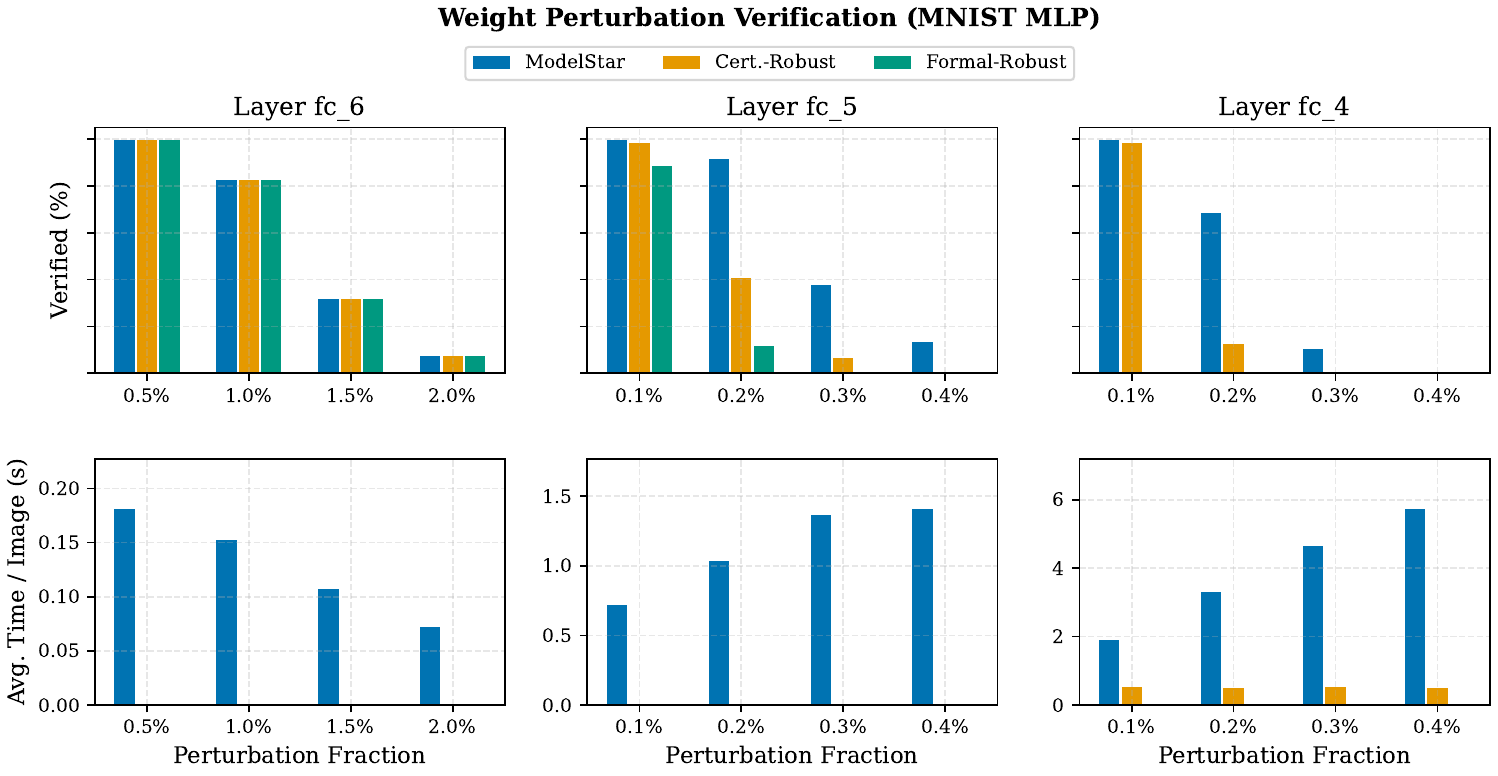}
\caption{Single-layer weight-perturbation verification on the MNIST MLP. (Top) fraction of images verified safe; (bottom) average execution time per image, both vs.\ the $L_\infty$-norm perturbation magnitude.}
\label{fig:weight-pert-MNIST-MLP}
\end{figure}

\textbf{Verification Under Weight and Parameter Perturbations.}
We evaluate ModelStar~\cite{zubair2025verification,zubair2026modelstar} on an MNIST MLP (5 hidden layers: 1024, 512, 256, 256, 256) against Certificated-Robust~\cite{weng2020towards} and Formal-Robust~\cite{tsai2021formalizing} (Fig.~\ref{fig:weight-pert-MNIST-MLP}). Perturbations are $L_\infty$ bounds on each layer's weight range, with magnitudes 0.05\%/0.1\%/0.2\%/0.4\% corresponding to 10-/9-/8-/7-bit quantization rounding error.
While ModelStar supports independently varying perturbations for individual weights \cite{zubair2026modelstar}, the baselines support only uniform row-wise~\cite{weng2020towards} or matrix-wise~\cite{tsai2021formalizing} perturbations. To ensure a fair comparison, we use a common perturbation magnitude for all weights in each perturbed layer and evaluate one layer's robustness against one perturbation magnitude at a time.
On 100 MNIST test images, ModelStar consistently matches or exceeds prior bounds: at 0.2\% perturbation on \texttt{fc\_4}, it verifies the safe classification of 69/100 images versus 13 for Certificated-Robust (an absolute gain of 56 percentage points).
For all three approaches, the classification safety of the NN for the unverified images is unknown due to over-approximation.
The scalability of ModelStar is limited by the width and number of perturbed layers: perturbing wide or multiple layers substantially increases Star-set dimensionality, leading to higher LP-solving times in subsequent nonlinear layers.
These results demonstrate that the ModelStar extension allows NNV3 to certify network robustness under quantization or hardware-induced weight uncertainty, addressing a deployment threat model beyond the reach of input-only verifiers.

\begin{figure}[t]
\centering
\includegraphics[width=0.7\linewidth]{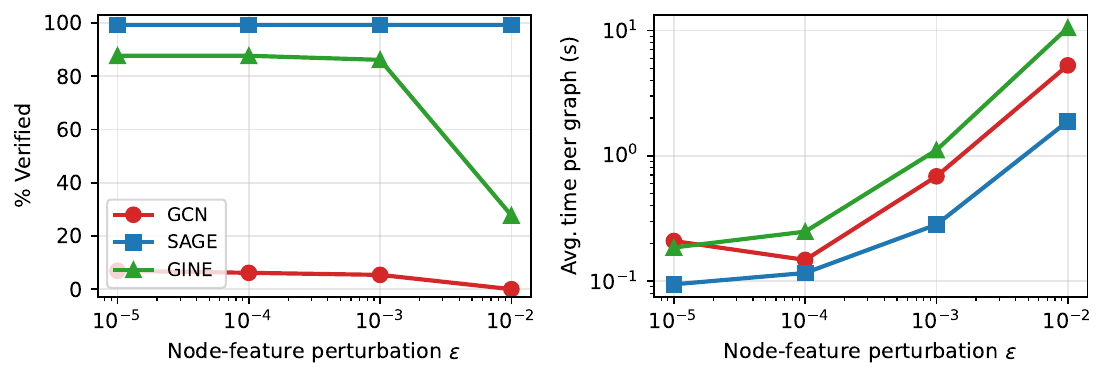}
\caption{GNNV verification on IEEE-24 power flow across three architectures (GCN, SAGE, GINE-Conv) vs.\ node-feature perturbation $\epsilon$ (log axis). (Left) percentage of voltage-magnitude nodes verified. (Right) average verification time per graph instance (log scale).}
\label{fig:gnnv_results}
\end{figure}

\textbf{Verification of Graph Neural Networks.}
We evaluate GraphStar~\cite{tumlin2026reachability} on AC power flow (PF) for the IEEE-24 bus system using 10 graph instances each for GCN, SAGE, and GINE-Conv models under the ML4ACOPF perturbation scheme~\cite{vnncomp2025}. We consider $L_\infty$ perturbations to active and reactive power node features for $\epsilon \in \{10^{-5}, 10^{-4}, 10^{-3}, 10^{-2}\}$, with voltage magnitude as the safety constraint. A common specification and reachability configuration is applied across all three architectures, demonstrating GraphStar's ability to verify differing message-passing architectures under a consistent threat model. Figure~\ref{fig:gnnv_results} summarizes this evaluation by reporting the percentage of voltage-magnitude nodes verified and the average verification time per graph instance for each architecture; the remaining nodes correspond to either proven violations or unknown outcomes. As GraphStar's computational complexity depends on graph size and network depth, this evaluation serves as a feasibility demonstration on a representative power-grid system rather than a comprehensive scalability study. This demonstrates reachability-based verification of GNN surrogates and supports formal analysis of GNN-based estimators.

\textbf{Verification of Spatio-Temporal and Volumetric Data.}
We evaluate VolumeStar~\cite{sasaki2025videoclass} on the ZoomIn-4f benchmark, a 4-frame MNIST-video classifier, under $L_\infty$ perturbations $\epsilon \in \{1/255, 2/255, 3/255\}$ with a 30-minute timeout per sample. VolumeStar verifies 7/10 (70\%) of cases at every $\epsilon$ tested (Table~\ref{tab:videostar_results}); the remaining three are unknown due to over-approximation. The scalability of VolumeStar is limited by frame count and volume: additional frames and larger spatial dimensions enlarge the generator basis, increasing memory and reachability cost per sample. This delivers the first reachability-based robustness certification for 3D-convolutional video classifiers, a modality where ImageStar-based propagation is intractable.

\textbf{Probabilistic Verification.}
We evaluate \textsf{NNV3}'s probabilistic verification pipeline~\cite{hashemi2025scaling} on the TinyYOLO object detector from VNN-COMP 2023~\cite{brix2023fourthinternationalverificationneural}. The approach combines randomized falsification with conformal prediction–based reachability analysis. A surrogate model is trained to approximate the network, and conformal inference over a calibration set bounds its error, yielding a reachable set (Table~\ref{tab:probver_results}). The guarantee is two-level: with confidence at least $99.9\%$ over the draw of the calibration set, a fresh input from the sampling distribution has its output inside the reachable set with probability at least $99.9\%$, where the confidence follows from the calibration size and rank through a Beta tail bound~\cite{hashemi2025scaling}.

We evaluate three randomly selected \textsf{VNNLIB} property specifications from the benchmark's 72 instances. All three properties are verified as UNSAT (specification holds), with GPU-accelerated verification requiring 111--132\,s per property. As in the sound mode, UNSAT means that the reachable set does not intersect the unsafe region. The set is not an over-approximation, however, and may omit the outputs of some inputs, so the verdict holds with the coverage and confidence stated above. Since the result is an ordinary Star set, the same specification routine checks it as in sound reachability, and any pipeline containing a CP-Star step yields a probabilistic verdict with that step's coverage and confidence. The dominant costs are the calibration size, which grows with the requested coverage and confidence, and property complexity, since the memory of the inflated set grows with the output dimension and every output constraint requires an LP over it. The evaluation therefore serves as a feasibility demonstration on a perception-scale model rather than a scalability study.

This extends \textsf{NNV} to instances for which sound reachability is intractable---for example, when the network is too large for the analysis to complete---since the probabilistic mode's cost scales with inference rather than with the number of unstable neurons.

\begin{figure}[t]
\centering
\begin{minipage}[b]{0.49\linewidth}
  \centering
  \captionof{table}{VolumeStar verification on ZoomIn-4f under $L_\infty$ perturbations (10 samples per $\epsilon$, 30-min timeout). \textbf{Ver.}~verified robust; \textbf{Unk.}~unknown due to over-approximation.}
  \label{tab:videostar_results}
  \small
  \setlength{\tabcolsep}{6pt}
  \begin{tabular}{cccc}
  \toprule
  $\epsilon$ & Ver. & Unk. & Avg.\ Time (s) \\
  \midrule
  $1/255$ & 7 & 3 & 35.54 \\
  $2/255$ & 7 & 3 & 37.88 \\
  $3/255$ & 7 & 3 & 36.49 \\
  \bottomrule
  \end{tabular}
\end{minipage}\hfill
\begin{minipage}[b]{0.49\linewidth}
  \centering
  \captionof{table}{Probabilistic verification on TinyYOLO (CP-Star, local Docker build with GPU). Coverage and confidence of $0.999$ each require $m = 9{,}230$ calibration samples per property.}
  \label{tab:probver_results}
  \small
  \setlength{\tabcolsep}{4pt}
  \begin{tabular}{cccc}
  \toprule
  Property & $\epsilon$ & Time (s) & Result \\
  \midrule
  Prop 101 & $1/255$ & 131.86 & UNSAT \\
  Prop 277 & $1/255$ & 111.92 & UNSAT \\
  Prop 356 & $1/255$ & 111.26 & UNSAT \\
  \bottomrule
  \end{tabular}
\end{minipage}
\end{figure}

\begin{table}[t]
\centering
\caption{FairNNV verification on Adult Census: Verified Fairness (VF, \%) and per-sample verification time (s). \emph{Counterfactual fairness} (CF) perturbs only the sensitive attribute ($\epsilon{=}0$); \emph{individual fairness} (IF) additionally perturbs non-sensitive features at radius $\epsilon$.}
\label{tab:fairness_results}
\small
\setlength{\tabcolsep}{7pt}
\begin{tabular}{l l c | c c c c c c}
\toprule
 & & CF & \multicolumn{6}{c}{IF ($\epsilon$)} \\
Model & metric & ($\epsilon{=}0$) & 0.01 & 0.02 & 0.03 & 0.05 & 0.07 & 0.10 \\
\midrule
Small  & VF (\%)   & 89    & 87    & 84    & 81    & 69    & 50    & 22 \\
       & Time (s)  & 0.78 & 0.89 & 1.06 & 1.40 & 2.17 & 3.41 & 5.53 \\
\midrule
Medium & VF (\%)   & 87    & 86    & 84    & 82    & 71    & 50    & 27 \\
       & Time (s)  & 0.72 & 2.64 & 5.76 & 10.10 & 21.75 & 39.73 & 98.70 \\
\bottomrule
\end{tabular}
\end{table}

\textbf{Fairness Verification.}
We evaluate FairNNV~\cite{tumlin2024fairnnv} on the Adult Census dataset~\cite{adult_census} using two FFNN classifiers: Small, with two hidden layers of 16 and 8 neurons, and Medium, with a single hidden layer of 50 neurons. Under counterfactual fairness, both reach Verified Fairness (VF) $\in [87,89]$\% with average verification time of 0.72--0.78\,s per sample; under individual fairness, VF degrades as $\epsilon$ grows, and the small model shows the steeper VF decline, while the medium model incurs substantially higher verification cost (Table~\ref{tab:fairness_results}). This delivers per-sample fairness certificates over continuous input regions. A direct comparison with FairSquare~\cite{albarghouthi2017fairsquare}, Justicia~\cite{ghosh2021justicia}, and abstract-interpretation-based fairness certifiers~\cite{urban2020perfectly} is not currently applicable because these approaches primarily target probabilistic or group fairness, whereas FairNNV certifies individual and counterfactual fairness through set-based reachability; extending FairNNV to group-fairness specifications is planned as future work and will enable more direct comparisons with these approaches.

\begin{table}[!t]
\centering
\footnotesize

\caption{Tool comparison on fully-connected VNNLIB benchmarks. Per cell: \textbf{V}~verified, \textbf{X}~violated, \textbf{U}~unknown, \textbf{T}~timeout, \textbf{S}~mean per-instance time (seconds). Timeout cap: 900\,s. AIVL uses \texttt{est-bnds}; \texttt{r-50} is NNV3's \texttt{relax-star-range-50}. Bold marks NNV3's strongest verification count per benchmark.}
\label{tab:aivl_fc}
\setlength{\tabcolsep}{2pt}
\renewcommand{\arraystretch}{1.05}
\newcolumntype{Y}{>{\centering\arraybackslash}p{1.6em}}
\newcolumntype{A}{>{\centering\arraybackslash}p{2.8em}}
\begin{tabular}{@{}l@{\hspace{3pt}} YYYYA @{\hspace{3pt}} YYYYA @{\hspace{3pt}} YYYYA@{}}
\toprule
 & \multicolumn{5}{c}{ACAS p3 ($N{=}20$)}
 & \multicolumn{5}{c}{ACAS p4 ($N{=}20$)}
 & \multicolumn{5}{c}{RL ($N{=}50$)} \\
\cmidrule(lr){2-6}\cmidrule(lr){7-11}\cmidrule(lr){12-16}
 & V & X & U & T & S
 & V & X & U & T & S
 & V & X & U & T & S \\
\midrule
AIVL   & 0 & 3 & 17 & 0 & 0.06
& 0 & 1 & 19 & 0 & 0.07
& 20 & 11 & 19 & 0 & 0.04 \\
\midrule
exact  & 8           & 3 & 0  & 9 & 58.23
& 9          & 3 & 0  & 8 & 83.99
& 32          & 15 & 1  & 2 & 5.36 \\
approx & \textbf{10} & 3 & 7  & 0 & 0.69 & 9 & 3 & 8  & 0 & 0.90
& 32 & 14 & 4  & 0 & 0.14 \\
r-50   & 1           & 2 & 17 & 0 & 0.30
& 1   & 2 & 17 & 0 & 0.30
& 32 & 14 & 4  & 0 & 0.08 \\
\bottomrule
\end{tabular}

\vspace{0.8em}

\captionof{table}{Tool comparison on convolutional VNNLIB benchmarks. Per cell: \textbf{V}~verified, \textbf{X}~violated, \textbf{U}~unknown, \textbf{T}~timeout, \textbf{S}~mean per-instance time (seconds). Timeout cap: 900\,s. AIVL uses \texttt{est-bnds}; NNV3 uses \texttt{approx-star}.}
\label{tab:aivl_cnn}
\setlength{\tabcolsep}{4pt}
\renewcommand{\arraystretch}{1.05}
\newcolumntype{Z}{>{\centering\arraybackslash}p{2em}}
\newcolumntype{B}{>{\centering\arraybackslash}p{3.4em}}
\begin{tabular}{@{}l@{\hspace{12pt}} ZZZZB @{\hspace{16pt}} ZZZZB@{}}
\toprule
 & \multicolumn{5}{c}{OVAL21 ($N{=}30$)}
 & \multicolumn{5}{c}{Collins RUL ($N{=}62$)} \\
\cmidrule(lr){2-6}\cmidrule(lr){7-11}
 & V & X & U & T & S
 & V & X & U & T & S \\
\midrule
AIVL   & 0 & 9           & 21 & 0 & 0.28 & 10          & 47 & 5 & 0 & 0.65 \\
\midrule
approx & 0 & 10 & 14 & 6 & 33.47 & 10 & 47 & 5 & 0 & 0.29 \\
\bottomrule
\end{tabular}

\vspace{0.8em}

\captionof{table}{MNIST-ResNet-8 robustness verification at perturbation magnitude $\varepsilon$ (25 test images per row). Both tools verify all 25 instances per row; the comparison reduces to runtime. AIVL uses \texttt{verifyNetworkRobustness} (DeepPoly, residual-network support added in R2024b); NNV3 uses \texttt{relax-star-area-50}.}
\label{tab:aivl_resnet}
\setlength{\tabcolsep}{8pt}
\renewcommand{\arraystretch}{1.05}
\begin{tabular}{@{}l rrrr@{}}
\toprule
Tool / $\varepsilon$ & $1/255$ & $2/255$ & $4/255$ & $8/255$ \\
\midrule
AIVL Time (s) & 10.29 & 10.32 & 10.28 & 10.41 \\
NNV3 Time (s) &  6.13 &  8.80 &  13.94 & 34.94 \\
\bottomrule
\end{tabular}

\end{table}

\textbf{Tool Comparison.}
We extend the NNV~2.0 head-to-head evaluation~\cite{lopez2023cavtool} against the MathWorks AI Verification Library (AIVL)~\cite{matlabDNNverification} using R2025b. We evaluate AIVL's interval-based output bounds for VNNLIB output specifications and its DeepPoly implementation for argmax robustness on residual architectures. The evaluation also incorporates two VNN-COMP-derived benchmarks, OVAL21~\cite{bak2021vnncomp} and Collins~RUL CNN~\cite{kirov2024collinsrul}, along with a natively trained MNIST-ResNet-8 (Tables~\ref{tab:aivl_fc}, \ref{tab:aivl_cnn}, \ref{tab:aivl_resnet}).

On ACAS Xu (Table~\ref{tab:aivl_fc}), AIVL's interval-based bounds, its only available option for half-space output specifications under R2025b, are insufficiently precise to verify any instances, reflecting the limited precision of the applicable verification method.\footnote{Under R2025b, \texttt{estimateNetworkOutputBounds} is AIVL's only documented path for half-space output specifications; the more precise \texttt{verifyNetworkRobustness} accepts argmax-robustness specifications only.} By contrast, \textsf{NNV3} verifies roughly half of the instances using either \texttt{exact-star} or \texttt{approx-star}, with \texttt{approx-star} completing every instance in sub-second time. On RL, \textsf{NNV3} verifies more instances than AIVL and identifies three additional counterexamples; the \texttt{relax-star-range-50} variant illustrates the precision--cost tradeoff, achieving sub-second runtime at a reduced verification rate. On the CNN benchmarks (Table~\ref{tab:aivl_cnn}), \texttt{approx-star} matches AIVL on Collins~RUL and identifies one additional counterexample on OVAL21. On MNIST-ResNet-8 (Table~\ref{tab:aivl_resnet}), both tools verify all $25$ instances at every $\epsilon$. \textsf{NNV3}'s runtime increases with $\epsilon$, whereas AIVL's remains approximately constant; at smaller perturbation bounds, \textsf{NNV3} completes verification faster than AIVL. The additional architectures and threat models supported through \textsf{NNV3}'s extensions to the Star-set family (Table~\ref{tab:tool_comparison}) fall outside AIVL's current coverage.

\vspace{-0.3cm}
\section{Related Work}
\label{sec:related_work}
\vspace{-0.1cm}
\begin{table}[t]
\centering
\newcolumntype{C}{>{\centering\arraybackslash}p{1.2em}}
\setlength{\tabcolsep}{2pt}
\rowcolors{2}{gray!15}{white}
\begin{tabular}{lCCCCCCCCCCCCCCCCp{12pt}}
&
\rotatebox[origin=l]{45}{\textbf{NNV3}} &
\rotatebox[origin=l]{45}{\textbf{AIVL}~\cite{matlabDNNverification}} &
\rotatebox[origin=l]{45}{\textbf{$\alpha,\beta$-Crown}~\cite{xu2021fast,wang2021beta}} &
\rotatebox[origin=l]{45}{\textbf{CORA}~\cite{Althoff2015arch,althoff2018implementation,ladner2025formalverificationgraphconvolutional}} &
\rotatebox[origin=l]{45}{\textbf{FastBATLLNN}~\cite{ferlez2022fast}} &
\rotatebox[origin=l]{45}{\textbf{JuliaReach}~\cite{bogomolov2019Juliareach,Schilling2022juliareachNNCS}} &
\rotatebox[origin=l]{45}{\textbf{Marabou}~\cite{katz2019marabou,wu2024marabou,wu2022jobscheduler}} &
\rotatebox[origin=l]{45}{\textbf{NeuralSAT}~\cite{duong2024neuralsat,duong2025neuralsat}} &
\rotatebox[origin=l]{45}{\textbf{NeVer2}~\cite{demarchi2024never2}} &
\rotatebox[origin=l]{45}{\textbf{nnenum}~\cite{bak2021nnenum}} &
\rotatebox[origin=l]{45}{\textbf{ReachNN}~\cite{huang2019reachnn,fan2020reachNN*}} &
\rotatebox[origin=l]{45}{\textbf{Reluplex}~\cite{katz2017reluplex}} &
\rotatebox[origin=l]{45}{\textbf{SobolBox}~\cite{das2025sobolbox}} & \rotatebox[origin=l]{45}{\textbf{StarV}~\cite{tran2025starv}} &
\rotatebox[origin=l]{45}{\textbf{Verinet}~\cite{henriksen2020verinet}} &
\rotatebox[origin=l]{45}{\textbf{Verisig}~\cite{Ivanov2019verisig}} &  \\
\midrule
\textbf{FFNN} & $\checkmark$ & $\checkmark$ & $\checkmark$ & $\checkmark$ & $\checkmark$ & $\checkmark$ & $\checkmark$ & $\checkmark$ & $\checkmark$ & $\checkmark$ & $\times$ & $\checkmark$ & $\checkmark$  & $\checkmark$ & $\checkmark$ & $\times$ & \\
\textbf{CNN} & $\checkmark$ & $\checkmark$ & $\checkmark$ & $\checkmark$ & $\times$ & $\checkmark*$ & $\checkmark$ & $\checkmark$ & $\checkmark$ & $\checkmark$ & $\times$ & $\times$ & $\checkmark$ & $\checkmark$ & $\checkmark$ & $\times$ & \\

\textbf{RNN} & $\checkmark$ & $\times$ & $\checkmark$ & $\times$ & $\times$ & $\times$ & $\times$ & $\times$ & $\times$ & $\times$ & $\times$ & $\times$ & $\times$  & $\checkmark$ & $\times$ & $\times$ & \\
\textbf{SSNN} & $\checkmark$ & $\times$ & $\checkmark^*$ & $\times$ & $\times$ & $\times$ & $\times$ & $\times$ & $\times$ & $\times$ & $\times$ & $\times$ & $\times$ & $\checkmark$ & $\times$ & $\times$ & \\
\textbf{\textit{\textcolor{purple}{TDNNs}}} & $\checkmark$ & $\times$ & $\times$ & $\times$ & $\times$ & $\times$ & $\times$ & $\times$ & $\times$ & $\times$ & $\times$ & $\times$ & $\times$ & $\times$ & $\times$ & $\times$ & \\
\textbf{\textit{\textcolor{purple}{GNNs}}} & $\checkmark$ & $\times$ & $\times$ & $\checkmark$ & $\times$ & $\times$ & $\checkmark^*$ & $\times$ & $\times$ & $\times$ & $\times$ & $\times$ & $\times$ & $\times$ & $\times$ & $\times$ & \\
\textbf{\textit{\textcolor{purple}{3D Data}}} & $\checkmark$ & $\times$ & $\times$ & $\times$ & $\times$ & $\times$ & $\times$ & $\times$ & $\times$ & $\times$ & $\times$ & $\times$ & $\times$ & $\times$ & $\times$ & $\times$ & \\
\textbf{\textit{\textcolor{purple}{WPP}}} & $\checkmark$ & $\times$ & $\times$ & $\times$ & $\times$ & $\times$ & $\times$ & $\times$ & $\times$ & $\times$ & $\times$ & $\times$ & $\times$ & $\times$ & $\times$ & $\times$ & \\
\textbf{NODEs} & $\checkmark$ & $\times$ & $\times$ & $\times$ & $\times$ & $\times$ & $\times$ & $\times$ & $\times$ & $\times$ & $\times$ & $\times$ & $\times$ & $\checkmark$ & $\times$ & $\times$ & \\
\textbf{NNCS} & $\checkmark$ & $\times$ & $\checkmark^*$ & $\checkmark$ & $\times$ & $\checkmark$ & $\times$ & $\times$ & $\times$ & $\times$ & $\checkmark$ & $\times$ & $\times$ & $\checkmark$ & $\times$  & $\checkmark$ & \\
\bottomrule
\end{tabular}
\vspace{0.1cm}
\caption{Comparison of neural network verification tools by supported architectures and applications, with $\checkmark$ meaning supported by a tool and $\times$ not supported. We use $^*$ to indicate partial support. In \textcolor{purple}{\textit{purple italics}}, the newly supported architecture/applications by \textsf{NNV3}. WPP signifies model perturbations.}
\vspace{-0.2cm}
\label{tab:tool_comparison}
\end{table}

Neural network verification has advanced through SMT-based methods (Reluplex~\cite{katz2017reluplex}, Marabou~\cite{katz2019marabou,wu2024marabou}, NeuralSAT~\cite{duong2025neuralsat}), bound propagation ($\alpha,\beta$-CROWN \cite{xu2021fast,wang2021beta}), abstract interpretation (AI2~\cite{gehr2018ai2}, DeepPoly~\cite{singh2019deeppoly}), and geometric approaches (Star sets~\cite{tran2019star}, zonotopes~\cite{singh2018fast}), among many other tools~\cite{brix2023first}.\footnote{See e.g., \url{https://vnn-comp.github.io/\#participants}}
These methods excel at verifying robustness of FFNNs and CNNs but typically focus on $L_p$-bounded input perturbations. Table~\ref{tab:tool_comparison} summarizes how \textsf{NNV3}'s coverage compares to existing tools across architectures and application domains.

\textbf{Weight and Parameter Perturbations.}
Most prior work assumes fixed network weights and verifies robustness only against input perturbations. Weight uncertainty from quantization, compression, or hardware faults has been studied via interval neural network (INN) verification~\cite{prabhakar2019abstraction}, interval bound propagation~\cite{weng2020towards}, and verification of quantized networks~\cite{henzinger2021scalable}. ModelStar~\cite{zubair2025verification,zubair2026modelstar} is complementary: INN intervals can be supplied as ModelStar perturbation specifications and propagated jointly with input uncertainty via star-set reachability.

\textbf{VolumeStar.}
Formal verification of video classifiers has previously been investigated only in~\cite{wu2020robustvideos}, using a game-theoretic framework over 2D convolutional and LSTM layers, with perturbations restricted to optical flow rather than video content. VolumeStar~\cite{sasaki2025videoclass} extends ImageStar~\cite{tran2020cav} from 2D images to 3D spatio-temporal data, supporting reachability through 3D convolutional and pooling layers under $L_\infty$ perturbations of the video content itself.

\textbf{Graph Neural Network Verification.}
Early GNN robustness work focused on empirical or probabilistic guarantees without formal soundness~\cite{wang2021gnnstructuralperturbation,tao2021immunization,lai2023towardCR}. Recent deterministic approaches include encoding message passing as feedforward networks for Marabou~\cite{wu2022jobscheduler}, matrix-polynomial-zonotope GCN reachability in CORA~\cite{ladner2025formalverificationgraphconvolutional}, and exact methods based on incremental constraint solving~\cite{liu_exact_2025} or MILP~\cite{hojny_verifying_2024}. These target node-classification on a narrow set of architectures; GraphStar~\cite{tumlin2026reachability} generalizes to edge-aware GNNs with joint node/edge uncertainty.

\textbf{Probabilistic Verification.}
Probabilistic verification offers a middle ground between exact methods (guarantees but limited scalability) and testing (scalability but no guarantees). Prior approaches include statistical sampling~\cite{webb2019statistical}, importance sampling for volume estimation~\cite{baluta2019quantitative}, randomized smoothing for classification~\cite{cohen2019certified} and segmentation~\cite{fischer2021scalable,hao2022gsmooth}, and conformal inference for classification~\cite{gendler2022adversarially,yan2024provably}. \textsf{NNV3} integrates conformal inference with set-based reachability, extending coverage guarantees to segmentation~\cite{hashemi2025scaling}.

\textbf{Fairness Verification.}
Fairness verification has been approached via probabilistic methods~\cite{albarghouthi2017fairsquare}, abstract interpretation~\cite{urban2020perfectly}, and SMT~\cite{ghosh2021justicia}. These techniques differ in the guarantee they provide: probabilistic methods yield statistical bounds, abstract interpretation produces conservative over-approximations, and SMT-based methods give satisfiability-style certificates. FairNNV~\cite{tumlin2024fairnnv} provides exact fairness certificates via reachability over continuous input regions.

\vspace{-0.3cm}
\section{Conclusion}
\label{sec:conclusion}
\vspace{-0.1cm}
\textsf{NNV3} advances neural network verification beyond architectural coverage to address new property classes (fairness), data modalities (video, volumetric), and practical deployment concerns (weight perturbations, probabilistic guarantees). While specialized tools may outperform \textsf{NNV} on specific benchmarks, \textsf{NNV} uniquely supports the broadest range of architectures (GNNs, TDNNs, 3D CNNs), data modalities (images, video, time series), and property classes (robustness, fairness, probabilistic guarantees, weight perturbations).

\textbf{Limitations and future work.} \textsf{NNV3} does not yet handle attention or autoregressive architectures, and ModelStar models continuous weight perturbations only, not fixed-point inference arithmetic. Ongoing work targets (i) a Python version to broaden adoption beyond the \textsf{MATLAB} ecosystem~\cite{sasaki2026n2v}, (ii) extending the Star-set family to transformers (work has already started), autoencoders, and temporal graph neural networks, (iii) discrete fixed-point arithmetic for ModelStar so that quantized inference can be verified end-to-end, and (iv) group-fairness definitions in FairNNV, which would enable direct comparison with probabilistic and group-fairness certifiers such as FairSquare~\cite{albarghouthi2017fairsquare} and Justicia~\cite{ghosh2021justicia}.

\textbf{Artifact availability.} The evaluation artifact accompanying this paper (models, specifications, and the scripts that reproduce the results reported in Section~\ref{sec:evaluation}) is archived on Zenodo at \url{https://doi.org/10.5281/zenodo.20433720} (release \texttt{v3.0-atva26}). \textsf{NNV3} is developed openly at \url{https://github.com/verivital/nnv}, with documentation, tutorials, and the developer API reference available at \url{https://verivital.github.io/nnv/}.

\subsubsection*{Acknowledgments.}
The material presented in this paper is based upon work supported by the National Science Foundation (NSF) through grant numbers 2220401 and 2325416, the Defense Advanced Research Projects Agency (DARPA) under contract number FA8750-23-C-0518, and the U.S. Department of Energy, Office of Science, Office of Advanced Scientific Computing Research, under Award Number DE-SC0025528. Any opinions, findings, and conclusions or recommendations expressed in this paper are those of the authors and do not necessarily reflect the views of DARPA, DOE, or NSF.

\bibliographystyle{splncs04}
\bibliography{bib/unified}

@misc{adult_census,
  author       = {Becker, Barry and Kohavi, Ronny},
  title        = {{Adult}},
  year         = {1996},
  howpublished = {UCI Machine Learning Repository},
  url         = {https://doi.org/10.24432/C5XW20}
}

@article{mehrabi2021survery,
author = {Mehrabi, Ninareh and Morstatter, Fred and Saxena, Nripsuta and Lerman, Kristina and Galstyan, Aram},
title = {A Survey on Bias and Fairness in Machine Learning},
year = {2021},
issue_date = {July 2022},
publisher = {Association for Computing Machinery},
address = {New York, NY, USA},
volume = {54},
number = {6},
issn = {0360-0300},
doi = {10.1145/3457607},
journal = {ACM Comput. Surv.},
month = jul,
articleno = {115},
numpages = {35}
}

@inproceedings{tumlin2024fairnnv,
author = {Tumlin, Anne M and Manzanas Lopez, Diego and Robinette, Preston and Zhao, Yuying and Derr, Tyler and Johnson, Taylor T},
title = {FairNNV: The Neural Network Verification Tool For Certifying Fairness},
year = {2024},
isbn = {9798400710810},
publisher = {Association for Computing Machinery},
address = {New York, NY, USA},
doi = {10.1145/3677052.3698677},
pages = {36–44},
numpages = {9},
location = {Brooklyn, NY, USA},
series = {ICAIF '24}
}

@inproceedings{kusner2017counterfactual,
author = {Kusner, Matt and Loftus, Joshua and Russell, Chris and Silva, Ricardo},
title = {Counterfactual fairness},
year = {2017},
isbn = {9781510860964},
publisher = {Curran Associates Inc.},
address = {Red Hook, NY, USA},
booktitle = {Proceedings of the 31st International Conference on Neural Information Processing Systems},
pages = {4069–4079},
numpages = {11},
location = {Long Beach, California, USA},
series = {NIPS'17}
}

@ARTICLE{wu2021gnnsurvey,
  author={Wu, Zonghan and Pan, Shirui and Chen, Fengwen and Long, Guodong and Zhang, Chengqi and Yu, Philip S.},
  journal={IEEE Transactions on Neural Networks and Learning Systems}, 
  title={A Comprehensive Survey on Graph Neural Networks}, 
  year={2021},
  volume={32},
  number={1},
  pages={4-24},
  doi={10.1109/TNNLS.2020.2978386}}

@inproceedings{
kipf2017semisupervised,
title={Semi-Supervised Classification with Graph Convolutional Networks},
author={Thomas N. Kipf and Max Welling},
booktitle={International Conference on Learning Representations},
year={2017},
url={https://openreview.net/forum?id=SJU4ayYgl}
}

@INPROCEEDINGS{sasaki2025videoclass,
  author={Sasaki, Samuel and Lopez, Diego Manzanas and Robinette, Preston K. and Johnson, Taylor T.},
  booktitle={2025 IEEE/ACM 13th International Conference on Formal Methods in Software Engineering (FormaliSE)}, 
  title={Robustness Verification of Video Classification Neural Networks}, 
  year={2025},
  volume={},
  number={},
  pages={22-33},
  doi={10.1109/FormaliSE66629.2025.00009}}

@ARTICLE{jaeger2014lung,
  author={Jaeger, Stefan and Karargyris, Alexandros and Candemir, Sema and Folio, Les and Siegelman, Jenifer and Callaghan, Fiona and Xue, Zhiyun and Palaniappan, Kannappan and Singh, Rahul K. and Antani, Sameer and Thoma, George and Wang, Yi-Xiang and Lu, Pu-Xuan and McDonald, Clement J.},
  journal={IEEE Transactions on Medical Imaging}, 
  title={Automatic Tuberculosis Screening Using Chest Radiographs}, 
  year={2014},
  volume={33},
  number={2},
  pages={233-245},
  doi={10.1109/TMI.2013.2284099}}

@article{candemir2014lung,
   author={Candemir, Sema and Jaeger, Stefan and Palaniappan, Kannappan and Musco, Jonathan P. and Singh, Rahul K. and Xue, Zhiyun and Karargyris, Alexandros and Antani, Sameer and Thoma, George and McDonald, Clement J.},
  journal={IEEE Transactions on Medical Imaging}, 
  title={Lung Segmentation in Chest Radiographs Using Anatomical Atlases With Nonrigid Registration}, 
  year={2014},
  volume={33},
  number={2},
  pages={577-590},
  doi={10.1109/TMI.2013.2290491}}

@misc{german_credit,
  author       = {Hofmann, Hans},
  title        = {{Statlog (German Credit Data)}},
  year         = {1994},
  howpublished = {UCI Machine Learning Repository},
  url         = {https://doi.org/10.24432/C5NC77}
}

@misc{bank_marketing,
  author       = {Moro, S. and Rita, P. and Cortez, P.},
  title        = {{Bank Marketing}},
  year         = {2014},
  howpublished = {UCI Machine Learning Repository},
  doi = {https://doi.org/10.24432/C5K306}
}

@inproceedings{tao2021immunization,
author = {Tao, Shuchang and Shen, Huawei and Cao, Qi and Hou, Liang and Cheng, Xueqi},
title = {Adversarial Immunization for Certifiable Robustness on Graphs},
year = {2021},
isbn = {9781450382977},
publisher = {Association for Computing Machinery},
address = {New York, NY, USA},
doi = {10.1145/3437963.3441782},
booktitle = {Proceedings of the 14th ACM International Conference on Web Search and Data Mining},
pages = {698–706},
numpages = {9},
location = {Virtual Event, Israel},
series = {WSDM '21}
}

@article{ladner2025formalverificationgraphconvolutional,
title={Formal Verification of Graph Convolutional Networks with Uncertain Node Features and Uncertain Graph Structure},
author={Tobias Ladner and Michael Eichelbeck and Matthias Althoff},
journal={Transactions on Machine Learning Research},
issn={2835-8856},
year={2025},
url={https://openreview.net/forum?id=B6y12Ot0cP},
note={}
}

@inproceedings{wang2021gnnstructuralperturbation,
author = {Wang, Binghui and Jia, Jinyuan and Cao, Xiaoyu and Gong, Neil Zhenqiang},
title = {Certified Robustness of Graph Neural Networks against Adversarial Structural Perturbation},
year = {2021},
isbn = {9781450383325},
publisher = {Association for Computing Machinery},
address = {New York, NY, USA},
doi = {10.1145/3447548.3467295},
booktitle = {Proceedings of the 27th ACM SIGKDD Conference on Knowledge Discovery \& Data Mining},
pages = {1645–1653},
numpages = {9},
location = {Virtual Event, Singapore},
series = {KDD '21}
}

@ARTICLE{lai2023towardCR,
  author={Lai, Yuni and Zhou, Jialong and Zhang, Xiaoge and Zhou, Kai},
  journal={IEEE Internet of Things Journal}, 
  title={Toward Certified Robustness of Graph Neural Networks in Adversarial AIoT Environments}, 
  year={2023},
  volume={10},
  number={15},
  pages={13920-13932},
  doi={10.1109/JIOT.2023.3263384}}

@article{wu2022jobscheduler,
author = {Wu, Haoze and Barrett, Clark and Sharif, Mahmood and Narodytska, Nina and Singh, Gagandeep},
title = {Scalable verification of GNN-based job schedulers},
year = {2022},
issue_date = {October 2022},
publisher = {Association for Computing Machinery},
address = {New York, NY, USA},
volume = {6},
number = {OOPSLA2},
doi = {10.1145/3563325},
journal = {Proc. ACM Program. Lang.},
month = oct,
articleno = {162},
numpages = {30}
}

@inproceedings{bak2021nnenum,
author = {Bak, Stanley},
title = {nnenum: Verification of ReLU Neural Networks with Optimized Abstraction Refinement},
year = {2021},
isbn = {978-3-030-76383-1},
publisher = {Springer-Verlag},
address = {Berlin, Heidelberg},
doi = {10.1007/978-3-030-76384-8\_2},
booktitle = {NASA Formal Methods: 13th International Symposium, NFM 2021, Virtual Event, May 24–28, 2021, Proceedings},
pages = {19–36},
numpages = {18}
}

@InProceedings{katz2019marabou,
author="Katz, Guy
and Huang, Derek A.
and Ibeling, Duligur
and Julian, Kyle
and Lazarus, Christopher
and Lim, Rachel
and Shah, Parth
and Thakoor, Shantanu
and Wu, Haoze
and Zelji{\'{c}}, Aleksandar
and Dill, David L.
and Kochenderfer, Mykel J.
and Barrett, Clark",
editor="Dillig, Isil
and Tasiran, Serdar",
title="The Marabou Framework for Verification and Analysis of Deep Neural Networks",
booktitle="Computer Aided Verification",
year="2019",
publisher="Springer International Publishing",
address="Cham",
pages="443--452",
isbn="978-3-030-25540-4"
}

@inproceedings{wang2021beta,
title={Beta-{CROWN}: Efficient Bound Propagation with Per-neuron Split Constraints for Neural Network Robustness Verification},
author={Shiqi Wang and Huan Zhang and Kaidi Xu and Xue Lin and Suman Jana and Cho-Jui Hsieh and J Zico Kolter},
booktitle={Advances in Neural Information Processing Systems},
editor={A. Beygelzimer and Y. Dauphin and P. Liang and J. Wortman Vaughan},
year={2021},
url={https://openreview.net/forum?id=ahYIlRBeCFw}
}

@inproceedings{sasaki2026arch,
  title={ARCH-COMP26 Category Report: Artificial Intelligence and Neural Network Control Systems (AINNCS) for Continuous and Hybrid Systems Plants},
  author={Sasaki, Samuel and Wooding, Ben and Johnson, Taylor T and Althoff, Matthias and Benet, Luis and Coogan, Samuel and Forets, Marcelo and Harapanahalli, Akash and Koller, Lukas and Ladner, Tobias and others},
  booktitle={Proceedings of 13th Int. Workshop on Applied Verification for Continuous and Hybrid Systems},
  volume={110},
  pages={85--130},
  year={2026}
}

@inproceedings{tumlin2026reachability,
  title={Reachability-Based Formal Verification of Graph Neural Networks with Node and Edge Features},
  author={Tumlin, Anne M and Wooding, Ben and Shao, Zhenxuan and Lopez, Diego Manzanas and Derr, Tyler and Johnson, Taylor T},
  booktitle={International Symposium on AI Verification},
  pages={271--298},
  year={2026},
  organization={Springer}
}

@inproceedings{ARCH_COMP25_Category_Report,
  author    = {Diego Manzanas Lopez and Matthias Althoff and Luis Benet and Samuel Coogan and Marcelo Forets and Akash Harapanahalli and Taylor T. Johnson and Tobias Ladner and Christian Schilling and Huan Zhang and Xiangru Zhong},
  title     = {ARCH-COMP25 Category Report: Artificial Intelligence and Neural Network Control Systems (AINNCS) for Continuous and Hybrid Systems Plants},
  booktitle = {Proceedings of 12th Int. Workshop on Applied Verification for Continuous and Hybrid Systems},
  editor    = {Goran Frehse and Matthias Althoff},
  series    = {EPiC Series in Computing},
  volume    = {108},
  publisher = {EasyChair},
  bibsource = {EasyChair, https://easychair.org},
  issn      = {2398-7340},
  url       = {/publications/paper/Gc39},
  doi       = {10.29007/9vg6},
  pages     = {71-121},
  year      = {2025}}

@inproceedings{hu2020strategies,
title={Strategies for Pre-training Graph Neural Networks},
author={Hu, Weihua and Liu, Bowen and Gomes, Joseph and Zitnik, Marinka and Liang, Percy and Pande, Vijay and Leskovec, Jure},
booktitle={International Conference on Learning Representations},
year={2020},
url={https://openreview.net/forum?id=HJlWWJSFDH},
}

@inproceedings{varbella2024powergraph,
author = {Varbella, Anna and Amara, Kenza and Gjorgiev, Blazhe and El-Assady, Mennatallah and Sansavini, Giovanni},
title = {PowerGraph: a power grid benchmark dataset for graph neural networks},
year = {2024},
isbn = {9798331314385},
publisher = {Curran Associates Inc.},
address = {Red Hook, NY, USA},
booktitle = {Proceedings of the 38th International Conference on Neural Information Processing Systems},
articleno = {3517},
numpages = {21},
location = {Vancouver, BC, Canada},
series = {NIPS '24}
}

@InProceedings{liu_exact_2025,
author="Liu, Minghao
and Lu, Chia-Hsuan
and Kwiatkowska, Marta",
editor="Sampaio, Augusto
and Stoelinga, Marielle",
title="Exact Verification of Graph Neural Networks with Incremental Constraint Solving",
booktitle="Formal Methods",
year="2026",
publisher="Springer Nature Switzerland",
address="Cham",
pages="641--662",
isbn="978-3-032-26204-2"
}

@inproceedings{hojny_verifying_2024,
	author = {Hojny, Christopher and Zhang, Shiqiang and Campos, Juan S. and Misener, Ruth},
    title = {Verifying message-passing neural networks via topology-based bounds tightening},
    year = {2024},
    publisher = {JMLR.org},
    booktitle = {Proceedings of the 41st International Conference on Machine Learning},
    articleno = {744},
    numpages = {26},
    location = {Vienna, Austria},
    series = {ICML'24}
}

@article{vnncomp2025,
   title={The 6th International Verification of Neural Networks Competition (VNN-COMP 2025): Summary and Results}, 
      author={Konstantin Kaulen and Tobias Ladner and Stanley Bak and Christopher Brix and Hai Duong and Thomas Flinkow and Taylor T. Johnson and Lukas Koller and Edoardo Manino and ThanhVu H Nguyen and Haoze Wu},
      year={2025},
      eprint={2512.19007},
      archivePrefix={arXiv},
      primaryClass={cs.LG},
      url={https://arxiv.org/abs/2512.19007}, 
}

@misc{brix2023fourthinternationalverificationneural,
      title={The Fourth International Verification of Neural Networks Competition (VNN-COMP 2023): Summary and Results}, 
      author={Christopher Brix and Stanley Bak and Changliu Liu and Taylor T. Johnson},
      year={2023},
      eprint={2312.16760},
      archivePrefix={arXiv},
      primaryClass={cs.LG},
      url={https://arxiv.org/abs/2312.16760}, 
}

@article{hashemi2025probabilistic,
    title={Probabilistic Robustness Analysis in High Dimensional Space: Application to Semantic Segmentation Network}, 
      author={Navid Hashemi and Samuel Sasaki and Diego Manzanas Lopez and Lars Lindemann and Ipek Oguz and Meiyi Ma and Taylor T. Johnson},
      year={2025},
      eprint={2509.11838},
      archivePrefix={arXiv},
      primaryClass={cs.CV},
      url={https://arxiv.org/abs/2509.11838}, 
}

@inproceedings{sasaki2026n2v,
  title={n2v: Neural Network Verification in Python (Competition Contribution)},
  author={Sasaki, Samuel and Wooding, Ben and Wang, Hanchen David and Tumlin, Anne M and Ma, Meiyi and Johnson, Taylor T},
  booktitle={International Symposium on AI Verification},
  pages={394--400},
  year={2026},
  organization={Springer}
}

@inproceedings{kirov2024collinsrul,
	address = {Cham},
	title = {Benchmark: {Remaining} {Useful} {Life} {Predictor} for {Aircraft} {Equipment}},
	isbn = {978-3-031-46002-9},
	booktitle = {Bridging the {Gap} {Between} {AI} and {Reality}},
	publisher = {Springer Nature Switzerland},
	author = {Kirov, Dmitrii and Rollini, Simone Fulvio},
	editor = {Steffen, Bernhard},
	year = {2024},
	pages = {299--304},
}

@article{zubair2026modelstar,
author = {Zubair, Muhammad Usama and Johnson, Taylor T. and Basu, Kanad and Abbas, Waseem},
title = {ModelStar: Reachability Analysis-based Safety Verification of Neural Networks Against Model Perturbations},
year = {2026},
issue_date = {May 2026},
publisher = {AI Access Foundation},
address = {El Segundo, CA, USA},
volume = {85},
issn = {1076-9757},
doi = {10.1613/jair.1.18922},
journal = {J. Artif. Int. Res.},
month = apr,
numpages = {29}
}

@inproceedings{Schilling2022juliareachNNCS,
  author    = {Christian Schilling and
               Marcelo Forets and
               Sebasti{\'{a}}n Guadalupe},
  title     = {{Verification of Neural-Network Control Systems by Integrating Taylor
               Models and Zonotopes}},
  booktitle = {{AAAI}},
  pages     = {8169--8177},
  publisher = {{AAAI} Press},
  year      = {2022},
  doi       = {10.1609/aaai.v36i7.20790}
}

@inbook{prabhakar2019abstraction,
author = {Prabhakar, Pavithra and Afzal, Zahra Rahimi},
title = {Abstraction based output range analysis for neural networks},
year = {2019},
publisher = {Curran Associates Inc.},
address = {Red Hook, NY, USA},
booktitle = {Proceedings of the 33rd International Conference on Neural Information Processing Systems},
articleno = {1414},
numpages = {11}
}

@inproceedings{Ivanov2019verisig,
author = {Ivanov, Radoslav and Weimer, James and Alur, Rajeev and Pappas, George J. and Lee, Insup},
title = {Verisig: verifying safety properties of hybrid systems with neural network controllers},
year = {2019},
isbn = {9781450362825},
publisher = {Association for Computing Machinery},
address = {New York, NY, USA},
doi = {10.1145/3302504.3311806},
booktitle = {Proceedings of the 22nd ACM International Conference on Hybrid Systems: Computation and Control},
pages = {169–178},
numpages = {10},
location = {Montreal, Quebec, Canada},
series = {HSCC '19}
}

@inproceedings{tran2020cavtool,
author = "Hoang-Dung Tran and Xiaodong Yang and Diego Manzanas Lopez and Patrick Musau and Luan Viet Nguyen and Weiming Xiang and Stanley Bak and Taylor T. Johnson",
title = "{NNV}: The Neural Network Verification Tool for Deep Neural Networks and Learning-Enabled Cyber-Physical Systems",
booktitle = "32nd International Conference on Computer-Aided Verification (CAV)",
year = "2020",
month = "July",
}

@Manual{onnx,
author = "{Open Neural Network Exchange (ONNX)}",
url = "https://github.com/onnx/onnx"
}

@Manual{matlabDNNverification,
    title = {AI Verification Library},
    author = {{The MathWorks, Inc.}},
    address = {Natick, Massachusetts, United States},
    year = {2025},
    url = {https://www.mathworks.com/products/ai-verification-library.html},
    version = {R2025b},
  }

@misc{vnnlib,
 title={VNN-LIB 2.0: Rigorous Foundations for Neural Network Verification}, 
      author={Ann Roy and Allen Antony and Andrea Gimelli and Matthew L. Daggitt},
      year={2026},
      eprint={2605.07451},
      archivePrefix={arXiv},
      primaryClass={cs.LG},
      url={https://arxiv.org/abs/2605.07451}, 
}

@inproceedings{bogomolov2019Juliareach,
  title={JuliaReach: a toolbox for set-based reachability},
  author={Bogomolov, Sergiy and Forets, Marcelo and Frehse, Goran and Potomkin, Kostiantyn and Schilling, Christian},
  booktitle={Proceedings of the 22nd ACM International Conference on Hybrid Systems: Computation and Control},
  pages={39--44},
  year={2019}
}

@InProceedings{fan2020reachNN*,
author="Fan, Jiameng
and Huang, Chao
and Chen, Xin
and Li, Wenchao
and Zhu, Qi",
editor="Hung, Dang Van
and Sokolsky, Oleg",
title={{ReachNN*: A Tool for Reachability Analysis of Neural-Network Controlled Systems}},
booktitle={{Automated Technology for Verification and Analysis}},
year="2020",
publisher="Springer International Publishing",
address="Cham",
pages="537--542",
isbn="978-3-030-59152-6"
}

@article{bak2021vnncomp,
title={The Second International Verification of Neural Networks Competition (VNN-COMP 2021): Summary and Results}, 
      author={Stanley Bak and Changliu Liu and Taylor Johnson},
      year={2021},
      eprint={2109.00498},
      archivePrefix={arXiv},
      primaryClass={cs.LO},
      url={https://arxiv.org/abs/2109.00498}, 
}

@inproceedings{xu2021fast,
title={Fast and Complete: Enabling Complete Neural Network Verification with Rapid and Massively Parallel Incomplete Verifiers},
author={Kaidi Xu and Huan Zhang and Shiqi Wang and Yihan Wang and Suman Jana and Xue Lin and Cho-Jui Hsieh},
booktitle={International Conference on Learning Representations},
year={2021},
url={https://openreview.net/forum?id=nVZtXBI6LNn}
}

@inproceedings{Althoff2015arch,
	author			= {Matthias Althoff},
	title			= {An Introduction to {CORA} 2015},
	booktitle		= {Proc. of the 1st and 2nd Workshop on Applied Verification for Continuous and Hybrid Systems},
	year			= {2015},
	publisher 		= {EasyChair},
	pages			= {120-151},
	doi		    	= {10.29007/zbkv}
  }

@inproceedings{singh2018fast,
author = {Singh, Gagandeep and Gehr, Timon and Mirman, Matthew and P\"{u}schel, Markus and Vechev, Martin},
title = {Fast and effective robustness certification},
year = {2018},
publisher = {Curran Associates Inc.},
address = {Red Hook, NY, USA},
booktitle = {Proceedings of the 32nd International Conference on Neural Information Processing Systems},
pages = {10825–10836},
numpages = {12},
location = {Montr\'{e}al, Canada},
series = {NIPS'18}
}

@inproceedings{gehr2018ai2,
  title={Ai2: Safety and robustness certification of neural networks with abstract interpretation},
  author={Gehr, Timon and Mirman, Matthew and Drachsler-Cohen, Dana and Tsankov, Petar and Chaudhuri, Swarat and Vechev, Martin},
  booktitle={2018 IEEE Symposium on Security and Privacy (SP)},
  pages={3--18},
  year={2018},
  organization={IEEE}
}

@article{brix2023first,
  title = {First three years of the international verification of neural networks competition ({VNN}-{COMP})},
	volume = {25},
	issn = {1433-2787},
	doi = {10.1007/s10009-023-00703-4},
	number = {3},
	journal = {International Journal on Software Tools for Technology Transfer},
	author = {Brix, Christopher and Müller, Mark Niklas and Bak, Stanley and Johnson, Taylor T. and Liu, Changliu},
	month = jun,
	year = {2023},
	pages = {329--339},
}

@inproceedings{wu2024marabou,
  title={{Marabou 2.0: A Versatile Formal Analyzer of Neural Networks}},
  author={Wu, Haoze and Isac, Omri and Zelji{\'c}, Aleksandar and Tagomori, Teruhiro and Daggitt, Matthew and Kokke, Wen and Refaeli, Idan and Amir, Guy and Julian, Kyle and Bassan, Shahaf and others},
  booktitle={Computer Aided Verification: 36th International Conference, CAV 2024},
  year={2024},
  organization={Springer}
}

@InProceedings{lopez2023cavtool,
author="Lopez, Diego Manzanas
and Choi, Sung Woo
and Tran, Hoang-Dung
and Johnson, Taylor T.",
editor="Enea, Constantin
and Lal, Akash",
title="NNV 2.0: The Neural Network Verification Tool",
booktitle="Computer Aided Verification",
year="2023",
publisher="Springer Nature Switzerland",
address="Cham",
pages="397--412",
isbn="978-3-031-37703-7"
}

@inproceedings{tran2021cav,
author = {Tran, Hoang-Dung and Pal, Neelanjana and Musau, Patrick and Lopez, Diego Manzanas and Hamilton, Nathaniel and Yang, Xiaodong and Bak, Stanley and Johnson, Taylor T.},
title = {Robustness Verification of Semantic Segmentation Neural Networks Using Relaxed Reachability},
year = {2021},
isbn = {978-3-030-81684-1},
publisher = {Springer-Verlag},
address = {Berlin, Heidelberg},
doi = {10.1007/978-3-030-81685-8\_12},
booktitle = {Computer Aided Verification: 33rd International Conference, CAV 2021, Virtual Event, July 20–23, 2021, Proceedings, Part I},
pages = {263–286},
numpages = {24}
}

@inproceedings{tran2015iccv,
  author={Tran, Du and Bourdev, Lubomir and Fergus, Rob and Torresani, Lorenzo and Paluri, Manohar},
  booktitle={2015 IEEE International Conference on Computer Vision (ICCV)}, 
  title={Learning Spatiotemporal Features with 3D Convolutional Networks}, 
  year={2015},
  volume={},
  number={},
  pages={4489-4497},
  doi={10.1109/ICCV.2015.510}}

@INPROCEEDINGS{carreira2017i3d,
  author={Carreira, João and Zisserman, Andrew},
  booktitle={2017 IEEE Conference on Computer Vision and Pattern Recognition (CVPR)}, 
  title={Quo Vadis, Action Recognition? A New Model and the Kinetics Dataset}, 
  year={2017},
  volume={},
  number={},
  pages={4724-4733},
  doi={10.1109/CVPR.2017.502}}

@INPROCEEDINGS{wu2020robustvideos,
  author={Wu, Min and Kwiatkowska, Marta},
  booktitle={2020 IEEE/CVF Conference on Computer Vision and Pattern Recognition (CVPR)}, 
  title={Robustness Guarantees for Deep Neural Networks on Videos}, 
  year={2020},
  volume={},
  number={},
  pages={308-317},
  doi={10.1109/CVPR42600.2020.00039}}

@misc{duong2024neuralsat,
      title={A DPLL(T) Framework for Verifying Deep Neural Networks}, 
      author={Hai Duong and ThanhVu Nguyen and Matthew Dwyer},
      year={2024},
      eprint={2307.10266},
      archivePrefix={arXiv},
      primaryClass={cs.LG},
      url={https://arxiv.org/abs/2307.10266}, 
}

@inproceedings{
hashemi2025scaling,
title={Scaling Data-Driven Probabilistic Robustness Analysis for Semantic Segmentation Neural Networks},
author={Navid Hashemi and Samuel Sasaki and Ipek Oguz and Meiyi Ma and Taylor T Johnson},
booktitle={The Thirty-ninth Annual Conference on Neural Information Processing Systems},
year={2025},
url={https://openreview.net/forum?id=liefJOFVfH}
}

@inproceedings{preston2024malware,
author = {Robinette, Preston K. and Manzanas Lopez, Diego and Serbinowska, Serena and Leach, Kevin and Johnson, Taylor T},
title = {Case Study: Neural Network Malware Detection Verification for Feature and Image Datasets},
year = {2024},
isbn = {9798400705892},
publisher = {Association for Computing Machinery},
address = {New York, NY, USA},
doi = {10.1145/3644033.3644372},
booktitle = {Proceedings of the 2024 IEEE/ACM 12th International Conference on Formal Methods in Software Engineering (FormaliSE)},
pages = {127–137},
numpages = {11},
location = {Lisbon, Portugal},
series = {FormaliSE '24}
}

@inproceedings{henriksen2020verinet,
  title={Efficient Neural Network Verification via Adaptive Refinement and Adversarial Search},
  author={Henriksen, Patrick and Lomuscio, Alessio},
  booktitle={European Conference on Artificial Intelligence (ECAI)},
  pages={2513--2520},
  year={2020}
}

@InProceedings{katz2017reluplex,
author="Katz, Guy
and Barrett, Clark
and Dill, David L.
and Julian, Kyle
and Kochenderfer, Mykel J.",
editor="Majumdar, Rupak
and Kun{\v{c}}ak, Viktor",
title="Reluplex: An Efficient SMT Solver for Verifying Deep Neural Networks",
booktitle="Computer Aided Verification",
year="2017",
publisher="Springer International Publishing",
address="Cham",
pages="97--117",
isbn="978-3-319-63387-9"
}

@article{singh2019deeppoly,
author = {Singh, Gagandeep and Gehr, Timon and P\"{u}schel, Markus and Vechev, Martin},
title = {An abstract domain for certifying neural networks},
year = {2019},
issue_date = {January 2019},
publisher = {Association for Computing Machinery},
address = {New York, NY, USA},
volume = {3},
number = {POPL},
doi = {10.1145/3290354},
journal = {Proc. ACM Program. Lang.},
month = jan,
articleno = {41},
numpages = {30}
}

@inproceedings{tran2019star,
author = {Tran, Hoang-Dung and Manzanas Lopez, Diago and Musau, Patrick and Yang, Xiaodong and Nguyen, Luan Viet and Xiang, Weiming and Johnson, Taylor T.},
title = {Star-Based Reachability Analysis of Deep Neural Networks},
year = {2019},
isbn = {978-3-030-30941-1},
publisher = {Springer-Verlag},
address = {Berlin, Heidelberg},
doi = {10.1007/978-3-030-30942-8\_39},
booktitle = {Formal Methods – The Next 30 Years: Third World Congress, FM 2019, Porto, Portugal, October 7–11, 2019, Proceedings},
pages = {670–686},
numpages = {17},
location = {Porto, Portugal}
}

@article{albarghouthi2017fairsquare,
author = {Albarghouthi, Aws and D'Antoni, Loris and Drews, Samuel and Nori, Aditya V.},
title = {FairSquare: probabilistic verification of program fairness},
year = {2017},
issue_date = {October 2017},
publisher = {Association for Computing Machinery},
address = {New York, NY, USA},
volume = {1},
number = {OOPSLA},
doi = {10.1145/3133904},
journal = {Proc. ACM Program. Lang.},
month = oct,
articleno = {80},
numpages = {30}
}

@article{urban2020perfectly,
author = {Urban, Caterina and Christakis, Maria and W\"{u}stholz, Valentin and Zhang, Fuyuan},
title = {Perfectly parallel fairness certification of neural networks},
year = {2020},
issue_date = {November 2020},
publisher = {Association for Computing Machinery},
address = {New York, NY, USA},
volume = {4},
number = {OOPSLA},
doi = {10.1145/3428253},
journal = {Proc. ACM Program. Lang.},
month = nov,
articleno = {185},
numpages = {30}
}

@inproceedings{ghosh2021justicia,
  	title = {Justicia: {A} {Stochastic} {SAT} {Approach} to {Formally} {Verify} {Fairness}},
	volume = {35},
	doi = {10.1609/aaai.v35i9.16925},
	number = {9},
	journal = {Proceedings of the AAAI Conference on Artificial Intelligence},
	author = {Ghosh, Bishwamittra and Basu, Debabrota and Meel, Kuldeep S.},
	month = may,
	year = {2021},
	pages = {7554--7563},
}

@inproceedings{henzinger2021scalable,
  title={Scalable Verification of Quantized Neural Networks},
  author={Henzinger, Thomas A. and Lechner, Mathias and Zikelic, Dorde},
  booktitle={AAAI Conference on Artificial Intelligence},
  pages={3787--3795},
  year={2021}
}

@inproceedings{
webb2019statistical,
title={Statistical Verification of Neural Networks},
author={Stefan Webb and Tom Rainforth and Yee Whye Teh and M. Pawan Kumar},
booktitle={International Conference on Learning Representations},
year={2019},
url={https://openreview.net/forum?id=S1xcx3C5FX},
}

@inproceedings{baluta2019quantitative,
author = {Baluta, Teodora and Shen, Shiqi and Shinde, Shweta and Meel, Kuldeep S. and Saxena, Prateek},
title = {Quantitative Verification of Neural Networks and Its Security Applications},
year = {2019},
isbn = {9781450367479},
publisher = {Association for Computing Machinery},
address = {New York, NY, USA},
doi = {10.1145/3319535.3354245},
booktitle = {Proceedings of the 2019 ACM SIGSAC Conference on Computer and Communications Security},
pages = {1249–1264},
numpages = {16},
location = {London, United Kingdom},
series = {CCS '19}
}

@inproceedings{cohen2019certified,
  title={Certified Adversarial Robustness via Randomized Smoothing},
  author={Cohen, Jeremy and Rosenfeld, Elan and Kolter, Zico},
  booktitle={International Conference on Machine Learning (ICML)},
  pages={1310--1320},
  year={2019}
}

@inproceedings{fischer2021scalable,
  title={Scalable Certified Segmentation via Randomized Smoothing},
  author={Fischer, Marc and Baader, Maximilian and Vechev, Martin},
  booktitle={International Conference on Machine Learning (ICML)},
  pages={3340--3351},
  year={2021}
}

@inproceedings{hao2022gsmooth,
  title={{GSmooth}: Certified Robustness against Semantic Transformations via Generalized Randomized Smoothing},
  author={Hao, Zhongkai and Ying, Chengyang and Su, Hang and Zhu, Jun and Song, Jian and Hu, Ze},
  booktitle={International Conference on Machine Learning (ICML)},
  pages={8465--8483},
  year={2022}
}

@inproceedings{gendler2022adversarially,
title={Adversarially Robust Conformal Prediction},
author={Asaf Gendler and Tsui-Wei Weng and Luca Daniel and Yaniv Romano},
booktitle={International Conference on Learning Representations},
year={2022},
url={https://openreview.net/forum?id=9L1BsI4wP1H}
}

@inproceedings{yan2024provably,
title={Provably Robust Conformal Prediction with Improved Efficiency},
author={Ge Yan and Yaniv Romano and Tsui-Wei Weng},
booktitle={The Twelfth International Conference on Learning Representations},
year={2024},
url={https://openreview.net/forum?id=BWAhEjXjeG}
}

@InProceedings{tran2020cav,
author="Tran, Hoang-Dung
and Bak, Stanley
and Xiang, Weiming
and Johnson, Taylor T.",
editor="Lahiri, Shuvendu K.
and Wang, Chao",
title="Verification of Deep Convolutional Neural Networks Using ImageStars",
booktitle="Computer Aided Verification",
year="2020",
publisher="Springer International Publishing",
address="Cham",
pages="18--42",
isbn="978-3-030-53288-8"
}

@InProceedings{pal2023fmics,
author="Pal, Neelanjana
and Lopez, Diego Manzanas
and Johnson, Taylor T.",
editor="Cimatti, Alessandro
and Titolo, Laura",
title="Robustness Verification of Deep Neural Networks Using Star-Based Reachability Analysis with Variable-Length Time Series Input",
booktitle="Formal Methods for Industrial Critical Systems",
year="2023",
publisher="Springer Nature Switzerland",
address="Cham",
pages="170--188",
isbn="978-3-031-43681-9"
}

@InProceedings{tran2019fm,
author="Tran, Hoang-Dung
and Musau, Patrick
and Diego Manzanas Lopez
and Xiaodong Yang
and Nguyen, Luan Viet
and Xiang, Weiming
and Taylor T. Johnson",
editor="",
title="Star-Based Reachability Analsysis for Deep Neural Networks",
booktitle="23rd International Symposisum on Formal Methods (FM'19)",
year="2019",
month="October",
publisher="Springer International Publishing",
}

@inproceedings{yan2022computing,
author = {Yan, Zheyu and Hu, Xiaobo Sharon and Shi, Yiyu},
title = {Computing-In-Memory Neural Network Accelerators for Safety-Critical Systems: Can Small Device Variations Be Disastrous?},
year = {2022},
isbn = {9781450392174},
publisher = {Association for Computing Machinery},
address = {New York, NY, USA},
doi = {10.1145/3508352.3549360},
articleno = {87},
numpages = {9},
location = {San Diego, California},
series = {ICCAD '22}
}

@inproceedings{zubair2025verification,
  author={Zubair, Muhammad Usama and Johnson, Taylor T. and Basu, Kanad and Abbas, Waseem},
  booktitle={2025 IEEE Conference on Artificial Intelligence (CAI)}, 
  title={Verification of Neural Network Robustness Against Weight Perturbations Using Star Sets}, 
  year={2025},
  volume={},
  number={},
  pages={637-642},
  doi={10.1109/CAI64502.2025.00117}
  }

@article{guo2018survey,
    title={A Survey on Methods and Theories of Quantized Neural Networks}, 
      author={Yunhui Guo},
      year={2018},
      eprint={1808.04752},
      archivePrefix={arXiv},
      primaryClass={cs.LG},
      url={https://arxiv.org/abs/1808.04752}, 
}

@inproceedings{sun2021exploring,
  title={Exploring the vulnerability of deep neural networks: A study of parameter corruption},
  author={Sun, Xu and Zhang, Zhiyuan and Ren, Xuancheng and Luo, Ruixuan and Li, Liangyou},
  booktitle={Proceedings of the AAAI Conference on Artificial Intelligence},
  volume={35},
  number={13},
  pages={11648--11656},
  year={2021}
}

@article{weng2020towards, 
title={Towards Certificated Model Robustness Against Weight Perturbations}, 
volume={34}, 
DOI={10.1609/aaai.v34i04.6105},
number={04}, 
journal={Proceedings of the AAAI Conference on Artificial Intelligence}, 
author={Weng, Tsui-Wei and Zhao, Pu and Liu, Sijia and Chen, Pin-Yu and Lin, Xue and Daniel, Luca}, 
year={2020}, 
month={Apr.}, 
pages={6356–6363} 
}

@inproceedings{tsai2021formalizing,
author = {Tsai, Yu-Lin and Hsu, Chia-Yi and Yu, Chia-Mu and Chen, Pin-Yu},
title = {Formalizing generalization and adversarial robustness of neural networks to weight perturbations},
year = {2021},
isbn = {9781713845393},
publisher = {Curran Associates Inc.},
address = {Red Hook, NY, USA},
booktitle = {Proceedings of the 35th International Conference on Neural Information Processing Systems},
articleno = {1506},
numpages = {13},
series = {NIPS '21}
}

@article{yousef2022holistic,
author = {Yousef, Rammah and Gupta, Gaurav and Yousef, Nabhan and Khari, Manju},
title = {A holistic overview of deep learning approach in medical imaging},
year = {2022},
issue_date = {Jun 2022},
publisher = {Springer-Verlag},
address = {Berlin, Heidelberg},
volume = {28},
number = {3},
issn = {0942-4962},
doi = {10.1007/s00530-021-00884-5},
journal = {Multimedia Syst.},
month = jun,
pages = {881–914},
numpages = {34}
}

@inproceedings{tran2023tutorial,
author = {Tran, Hoang-Dung and Manzanas Lopez, Diego and Johnson, Taylor},
title = {Tutorial: Neural Network and Autonomous Cyber-Physical Systems Formal Verification for Trustworthy AI and Safe Autonomy},
year = {2024},
isbn = {9798400702914},
publisher = {Association for Computing Machinery},
address = {New York, NY, USA},
doi = {10.1145/3607890.3608454},
booktitle = {Proceedings of the International Conference on Embedded Software},
pages = {1–2},
numpages = {2},
location = {Hamburg, Germany},
series = {EMSOFT '23}
}

@INPROCEEDINGS{johnson2024tutorial,
  author={Johnson, Taylor T. and Lopez, Diego Manzanas and Tran, Hoang-Dung},
  booktitle={2024 54th Annual IEEE/IFIP International Conference on Dependable Systems and Networks - Supplemental Volume (DSN-S)}, 
  title={Tutorial: Safe, Secure, and Trustworthy Artificial Intelligence (AI) via Formal Verification of Neural Networks and Autonomous Cyber-Physical Systems (CPS) with NNV}, 
  year={2024},
  volume={},
  number={},
  pages={65-66},
  doi={10.1109/DSN-S60304.2024.00027}}

@inproceedings{althoff2018implementation,
  author    = {Matthias Althoff and Dmitry Grebenyuk and Niklas Kochdumper},
  title     = {Implementation of Taylor models in CORA 2018},
  booktitle = {ARCH18. 5th International Workshop on Applied Verification of Continuous and Hybrid Systems},
  editor    = {Goran Frehse},
  series    = {EPiC Series in Computing},
  volume    = {54},
  publisher = {EasyChair},
  bibsource = {EasyChair, https://easychair.org},
  issn      = {2398-7340},
  doi       = {10.29007/zzc7},
  pages     = {145-173},
  year      = {2018}}

@inproceedings{ferlez2022fast,
  author = {Ferlez, James and Khedr, Haitham and Shoukry, Yasser},
title = {Fast BATLLNN: Fast Box Analysis of Two-Level Lattice Neural Networks},
year = {2022},
isbn = {9781450391962},
publisher = {Association for Computing Machinery},
address = {New York, NY, USA},
doi = {10.1145/3501710.3519533},
booktitle = {Proceedings of the 25th ACM International Conference on Hybrid Systems: Computation and Control},
articleno = {23},
numpages = {11},
location = {Milan, Italy},
series = {HSCC '22}
}

@inproceedings{duong2025neuralsat,
  title={NeuralSAT: a high-performance verification tool for deep neural networks},
  author={Duong, Hai and Nguyen, ThanhVu and Dwyer, Matthew B},
  booktitle={International Conference on Computer Aided Verification},
  pages={409--423},
  year={2025},
  organization={Springer},
  doi={10.1007/978-3-031-98679-6\_19}
}

@article{demarchi2024never2,
 title = {{NeVer2}: learning and verification of neural networks},
	volume = {28},
	issn = {1433-7479},
	doi = {10.1007/s00500-024-09907-5},
	number = {19},
	journal = {Soft Computing},
	author = {Demarchi, Stefano and Guidotti, Dario and Pulina, Luca and Tacchella, Armando},
	month = oct,
	year = {2024},
	pages = {11647--11665},
}

@inproceedings{das2025sobolbox,
  title={{SobolBox: Boxed Refinement of Sobol Sequence Samples for Neural Network Verification (Competition Contribution)}},
  author={Das, Sarthak},
  booktitle={International Symposium on AI Verification},
  pages={272--277},
  year={2025},
  organization={Springer}
}

@inproceedings{tran2020neural,
  author={Tran, Hoang-Dung and Lopez, Diego Manzanas and Yang, Xiaodong and Musau, Patrick and Nguyen, Luan Viet and Xiang, Weiming and Bak, Stanley and Johnson, Taylor T.},
  booktitle={2020 IEEE Workshop on Design Automation for CPS and IoT (DESTION)}, 
  title={Demo: The Neural Network Verification (NNV) Tool}, 
  year={2020},
  volume={},
  number={},
  pages={21-22},
  doi={10.1109/DESTION50928.2020.00010}}

@article{huang2019reachnn,
author = {Huang, Chao and Fan, Jiameng and Li, Wenchao and Chen, Xin and Zhu, Qi},
title = {ReachNN: Reachability Analysis of Neural-Network Controlled Systems},
year = {2019},
issue_date = {October 2019},
publisher = {Association for Computing Machinery},
address = {New York, NY, USA},
volume = {18},
number = {5s},
issn = {1539-9087},
doi = {10.1145/3358228},
journal = {ACM Trans. Embed. Comput. Syst.},
month = oct,
articleno = {106},
numpages = {22}
}

@inproceedings{tran2025starv,
author = {Tran, Hoang-Dung and Choi, Sung Woo and Li, Yuntao and Liu, Qing and Okamoto, Hideki and Hoxha, Bardh and Fainekos, Georgios},
title = {StarV: A Qualitative and Quantitative Verification Tool for Learning-Enabled Systems},
year = {2025},
isbn = {978-3-031-98678-9},
publisher = {Springer-Verlag},
address = {Berlin, Heidelberg},
doi = {10.1007/978-3-031-98679-6\_17},
booktitle = {Computer Aided Verification: 37th International Conference, CAV 2025, Zagreb, Croatia, July 23-25, 2025, Proceedings, Part II},
pages = {376–394},
numpages = {19},
location = {Zagreb, Croatia}
}

\end{document}